\documentclass[conference]{IEEEtran}
\usepackage{times}

\usepackage[numbers]{natbib}
\usepackage{multicol}
\usepackage[bookmarks=true]{hyperref}
\usepackage{booktabs}
\usepackage{microtype}
\usepackage{graphicx}
\usepackage{subcaption}
\usepackage{amsmath}
\usepackage{amssymb}
\usepackage{mathtools}
\usepackage{amsthm}
\usepackage{algorithm} 
\usepackage{algorithmic}
\usepackage{caption}
\let\labelindent\relax
\usepackage{enumitem}
\usepackage{multirow}
\usepackage{array}     
\usepackage{booktabs}  
\usepackage{makecell}  
\usepackage{comment}
\usepackage{textcomp}

\usepackage[textsize=tiny]{todonotes}
\usepackage{listings} 
\usepackage{xcolor}   
\usepackage{geometry} 
\newcommand{\our}{{\textit{VISTA}}}

\begin{document}
\makeatletter
\let\@oldmaketitle\@maketitle
\renewcommand{\@maketitle}{\@oldmaketitle
  \begin{center}
  \captionsetup{type=figure}
  \setcounter{figure}{0}
  \includegraphics[trim=0.4ex 0 0 0, clip, width=1.0\textwidth]{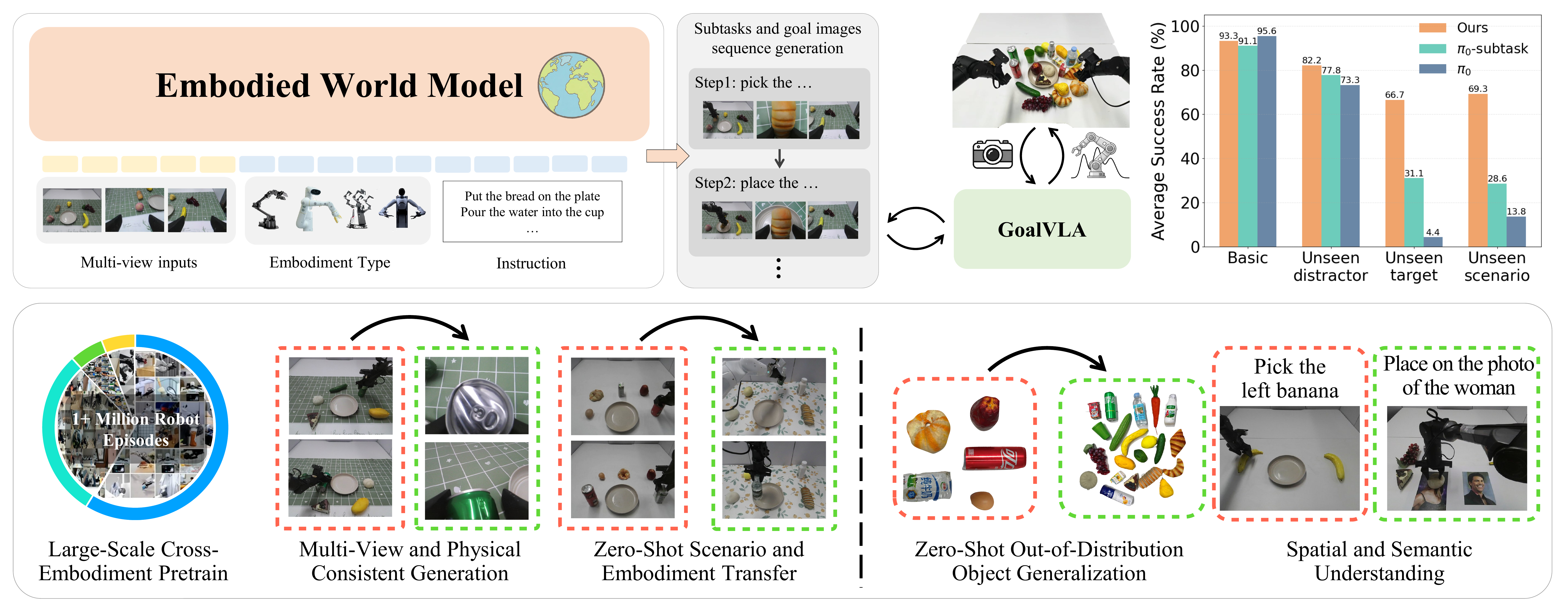}
    \caption{The illustration of our proposed hierarchical system \our{}. The world model is trained over massive cross-embodiment manipulation datasets and is capable of generating multi-view and physically consistent subtask sequences over novel scenarios, with textual subtask and visual goal images at each step. The subtask and goal images are further fed into the subtask and goal images conditioned VLA (GoalVLA) to provide detailed guidance. 
    With only 2 hours of real robot data collection on 5 objects, our approach could generalize to 21 unseen objects and even tasks that require spatial and semantic understanding.
    For the real-world evaluation, our approach stably maintains its performance over unseen target objects and novel scenarios, while the conventional VLA relying solely on language instruction guidance shows an obvious performance degradation, which demonstrates the data efficiency and generalization ability of our method.} 
    \vspace{-20pt}
    \label{fig:teaser}
  \end{center}
}
\makeatother

\title{Scaling World Model for\\ Hierarchical Manipulation Policies}

\author{
  Long Qian$^{\ast1,2}$, Yueze Wang$^{\ast2}$, Jiaxi Song$^{\ast2,3}$, Junbo Zhang$^{2,3}$, Peiyan Li$^{2,5}$, \\Wenxuan Wang$^{2,5}$, Yuqi Wang$^{2,5}$,
  Haoyang Li$^2$, Shaoxuan Xie$^2$, Guocai Yao$^2$, Hanbo Zhang$^4$, \\ Xinlong Wang$^2$, Zhongyuan Wang$^2$,
  Xuguang Lan$^{\dagger1}$, Huaping Liu$^{\dagger3}$, Xinghang Li$^{\dagger\ddag2}$ \\
  $^1$Xi'an Jiao Tong University,
  $^2$Beijing Academy of Artificial Intelligence,
  $^3$Tsinghua University, \\
  $^4$National University of Singapore,
  $^5$Institute of Automation, Chinese Academy of Sciences\\
$^\ast$Equal Contribution, $^\dagger$Corresponding Author, $^\ddag$Project Lead. \\
\texttt{qianlongym@stu.xjtu.edu\allowbreak.cn}, \texttt{xglan@mail.xjtu.edu.cn}, \\ \texttt{hpliu@tsinghua.edu.cn}, \texttt{lixingha23\allowbreak @mails.tsinghua.edu.cn}
}

\maketitle

\begingroup
\renewcommand{\thefootnote}{}  
\renewcommand{\footnoterule}{} 
\endgroup

\begin{abstract}

Vision-Language-Action (VLA) models are promising for generalist robot manipulation but remain brittle in out-of-distribution (OOD) settings, especially with limited real-robot data.
To resolve the generalization bottleneck, we introduce a hierarchical Vision-Language-Action framework \our{} that leverages the generalization of large-scale pre-trained world model for robust and generalizable \textbf{VI}sual \textbf{S}ubgoal \textbf{TA}sk decomposition (\our{}). 
Our hierarchical framework \our{} consists of a world model as the high-level planner and a VLA as the low-level executor.
The high-level world model first divides manipulation tasks into subtask sequences with goal images, and the low-level policy follows the textual and visual guidance to generate action sequences.
Compared to raw textual goal specification, these synthesized goal images provide visually and physically grounded details for low-level policies, making it feasible to generalize across unseen objects and novel scenarios. We validate both visual goal synthesis and our hierarchical VLA policies in massive out-of-distribution scenarios, and the performance of the same-structured VLA in novel scenarios could boost from 14\% to 69\% with the guidance generated by the world model. Results demonstrate that our method outperforms previous baselines with a clear margin, particularly in out-of-distribution scenarios. 
Project page: \href{https://vista-wm.github.io/}{https://vista-wm.github.io}

\end{abstract}

\IEEEpeerreviewmaketitle

\section{Introduction}

Vision-Language-Action (VLA) models~\citep{liu2025towards,bjorck2025gr00t,li2023vision} offer a promising paradigm for learning generalist robot manipulation policies by leveraging large-scale pretraining to directly follow natural language instructions and generate low-level actions. However, VLAs remain brittle in out-of-distribution (OOD) scenarios, particularly when real-world robot data is scarce. As analyzed in~\citep{hancock2025actions, grover2025enhancing}, this limitation stems from a fundamental mismatch in data structure between pretrained Vision-Language Models (VLMs) and VLAs. VLMs are trained on paired image–text data to model a discretized language distribution, whereas VLAs are trained on long robot trajectories consisting of hundreds of image–action pairs under a single instruction, with continuous action regression as the learning target. As a result, VLAs lack the zero-shot generalization ability of VLMs and require a large amount of real-world data for robust performance. 
This challenge is exacerbated in long-horizon manipulation, where the combinatorial growth of state transitions makes direct mapping from a static language command to precise action sequences ill-posed and highly data-intensive.

Hierarchical task decomposition is therefore essential. However, existing approaches face a representation trade-off: linguistic sub-goals generalize well semantically but lack concrete spatial and physical constraints~\citep{intelligence2025pi_, ye2025vla}, while dense video prediction provides richer detail but suffers from temporal drift and physical inconsistency over long horizons~\citep{chi2025wow, yang2025roboenvision}. This motivates a central question: \textit{How can we leverage foundation model generalization to abstract manipulation into an intermediate representation that improves both robustness and data efficiency?}


We answer this question with \our{}, a hierarchical Vision-Language-Action framework in which high-level planning is performed by a world model that decomposition manipulation into a sequence of compact, visually grounded subgoals, and low-level control is carried out by a VLA policy guided by these discretized visual subgoals. The key insight of our approach is that robust generalization can be achieved by reasoning over stable visual subgoals, rather than relying solely on brittle textual subtasks or dense, error-prone trajectory predictions.

Specifically, \our{} synthesizes a discrete sequence of goal images that serve as visual milestones, which are interleaved with textual subtasks to form goal-conditioned manipulation primitives. Predicting only key frames enables scalable training on large video datasets, mitigates hallucination through state invariance, and provides explicit spatial constraints that language alone cannot convey. Within \our{}, execution is handled by a goal-conditioned VLA policy that fuses the current observation with the synthesized goal image and textual instruction to predict action chunks. Trained at scale, this policy leverages high-fidelity visual goals as strong spatial priors, enabling precise execution and improved generalization to unseen objects and scenes.

We evaluate \our{} on multi-stage manipulation with an emphasis on out-of-distribution (OOD) generalization. Using only 2 hours of real-world teleoperation data collected on 5 objects, \our{} achieves 69\% success in novel scenarios spanning 21 unseen objects, outperforming the baseline \(\pi_0\) guided by language instruction, which only achieves 14\% success. These results highlight that our hierarchical generative visual decomposition substantially improves both data efficiency and robustness.

Our main contributions are as follows:
\begin{itemize}
\item A scalable data processing pipeline that relabels millions of robot trajectories into an interleaved format of textual subtasks and visual goal images.
\item A generative embodied world model that produces physically and multi-view consistent visual sub-goals for manipulation guidance.

\item The \textbf{VIsual Subgoal-based TAsk decomposition} framework \textbf{\our{}}, a hierarchical framework with a generative world model and a goal-image-conditioned policy (GoalVLA) that significantly outperforms standard baselines in OOD settings.
\end{itemize}

The remainder of the paper is organized as follows. Sec.~\ref{sec:rel_works} reviews related work, Sec.~\ref{sec:formulation} presents the problem formulation, and Sec.~\ref{sec:method} details \our{}. The experimental results are reported in Sec.~\ref{sec:exp}, followed by conclusions and limitations in Sec.~\ref{sec:conclusion} and Sec.~\ref{sec:limit}.

\begin{figure*}[!t]
    \centering
    \includegraphics[width=\textwidth]{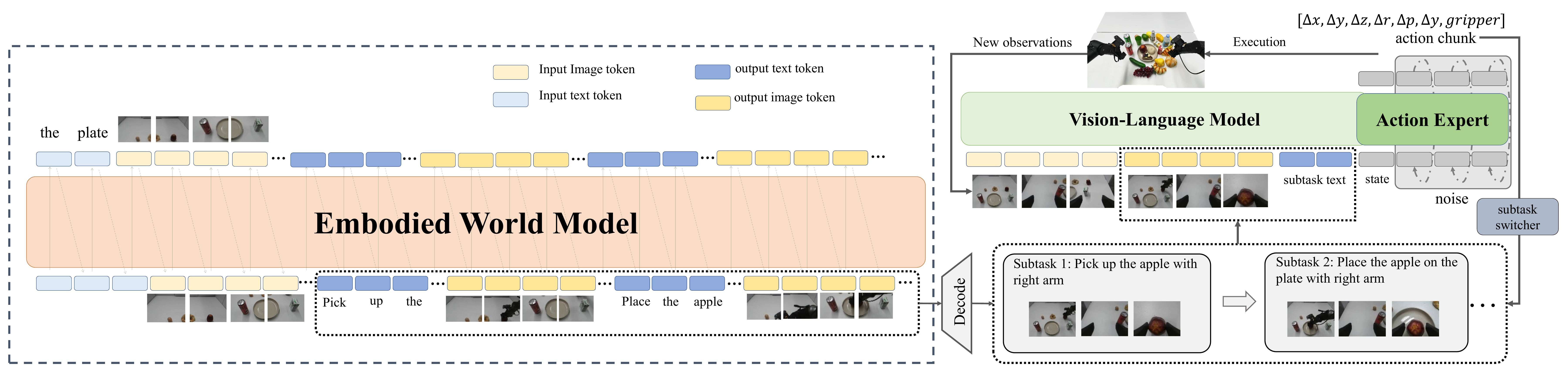}
    \vspace{-6pt}
    \caption{\textbf{Overview of the \our{}.} 
    The framework comprises two essential modules: 
    \textit{Left}: \our{} serves as a high-level planner. By treating visual goals and textual subtasks as a unified multi-modal sequence, it autoregressively generates interleaved textual subtasks and visual goals conditioned on the global instruction and initial observation. 
    \textit{Right}: The GoalVLA acts as the low-level controller. It takes the real-time observation and the generated subgoal as input to predict executable action chunks. The execution process is managed hierarchically, where a subtask switcher transitions to the next stage upon completion of the current stage.}
    \vspace{-15pt}
    \label{fig:method}
\end{figure*}

\section{Related Works}\label{sec:rel_works}
\subsection{Language-conditioned Imitation Learning}

Language-conditioned imitation learning typically aims to map visual observations and linguistic instructions directly to action sequences. Early works like BC-Z~\cite{jang2022bc} and CLIPort~\cite{shridhar2022cliport} relied on end-to-end behavior cloning with labeled data or generalizable semantic representations for modular design. Recently, the emergence of Vision-Language-Action (VLA) models~\cite{brohan2023rt1, zitkovich2023rt2, black2025pi0} fundamentally shifted this formulation by casting robot actions as tokens within a unified autoregressive sequence, enabling the direct transfer of semantic reasoning from internet-scale pre-training to robotic control. Yet, despite their promise as generalist policies, these models suffer from brittleness in generalization, particularly for long-horizon tasks with out-of-distribution objects. To address this, hierarchical decompositions such as Helix~\cite{figure2024helix} and G0~\cite{jiang2025galaxea} explicitly separate high-level planning from low-level execution. However, these methods typically rely on textual sub-goals, which lack the expressiveness required to enforce strict visual and physical constraints. Therefore, effectively grounding high-level semantics into physically consistent manipulation plans remains an open challenge that this work addresses.

\subsection{World Models for Robot Manipulation}

World models seek to internalize physical dynamics from large-scale unstructured data, enabling robots to predict action outcomes without costly real-world interaction. A prominent application is video pretraining, which leverages internet-scale human manipulation videos to learn physics and object affordances. Methods such as GR-1~\cite{wu2024gr1} cast this as a GPT-style sequence completion problem, learning strong visual representations by predicting future frames and treating robot actions as latent drivers of video evolution. Extending this paradigm, Inverse Dynamics~\cite{du2023learning} decouples high-level planning from low-level control by first generating future successful executions with a diffusion model, then decoding actions from the predicted futures. Beyond policy initialization, world models support rewards, planning, and decomposition. In RL, video generative models provide dense, differentiable rewards, as in VIPER~\cite{escontrela2023viper} and TeViR~\cite{chen2025tevir}, which score trajectories by their likelihood under pretrained video models. Model-based methods such as Dreamer~\cite{hafner2025dreamer} plan in learned latent spaces to achieve sample-efficient robot learning. To bridge passive video and control, latent action models like Genie~\cite{bruce2024genie}, LAPA~\cite{ye2025lapa}, and UniVLA~\cite{bu2025univla} learn discrete action codes via VQ-VAE objectives, effectively transforming videos into interactive simulators. In contrast, our work synthesizes sparse visual sub-goals as a temporal bottleneck, avoiding long-horizon video generation while providing explicit spatial constraints for robust execution.

\subsection{Manipulation Task Decomposition}

The decomposition of manipulation tasks aims to distill long-horizon behaviors into compact intermediate representations that are expressive enough for planning and structured enough for reliable execution. In the context of VLA systems, dual-system frameworks such as Helix~\cite{figure2024helix} and G0~\cite{jiang2025galaxea} implicitly adopt this decomposition principle by separating low-frequency semantic reasoning (e.g., VLM-based planning) from high-frequency motor control. More broadly, prior work has explored a spectrum of decomposition interfaces between language and action. Code-based decompositions (e.g., Code as Policies~\cite{liang2023codeaspolicy} and ProgPrompt~\cite{singh2023progprompt}) use executable programs as a structured interface for compositional reasoning, while neuro-symbolic approaches such as LLM+P~\cite{liu2023llmp} and SayPlan~\cite{rana2023sayplan} map natural language into PDDL-style plans to enforce logical consistency. Beyond purely linguistic decompositions, visual decompositions such as LEAP~\cite{nasiriany2019leap} and VLM-TDP~\cite{huang2025vlmtdp} represent tasks using latent embeddings or predicted spatial trajectories. These methods collectively reveal a central trade-off among decompositions expressiveness, interpretability, and verifiability. In contrast, we formulate task decompositions as generative visual subgoal synthesis, providing a lightweight yet grounded intermediate representation that preserves fine-grained spatial constraints while remaining easy to verify.

\section{Problem Formulation}\label{sec:formulation}

We define the language-conditioned manipulation problem in which a robot interacts with an environment to fulfill a global instruction $L\in\mathcal{T}$, where $\mathcal{T}$ is the free text space. 
The input of the system comprises the current visual observation $I_t\in\mathcal{I}$ at timestamp $t$, where $\mathcal{I}$ is the space of images, and the execution history. The output is a sequence of actions $\boldsymbol{a}\in\mathcal{A}^{k}$, i.e., an action chunk of length $k$, that drives the robot toward task completion. 

To address the challenge of compositional manipulation tasks, we do not map the global instruction directly to actions; instead, we decompose $L$ into a sequence of intermediate subtasks. The subtask at stage $i$ consists of a textual subtask $l_i\in\mathcal{T}$, specifying ``what to do'', and a visual goal $g_i\in\mathcal{I}$ specifying ``how to do it''. 
The objective is to train a policy $\pi$ that leverages these sub-goals to generate precise action trajectories.
Formally, we utilize a world model $W$ to predict the next required sub-goal based on the instruction $L$ and the history $\boldsymbol{h}=(l_0,g_0,...,l_{i-1},g_{i-1})$. The low-level policy $\pi_{\theta}$, parameterized by $\theta$, then takes the current observation $I_t$ and the current milestone of both the subtask and the visual goal ($l_i, g_i$) to infer the action sequence $\boldsymbol{a}$. This process is formulated as follows:
\begin{equation}
(s_i, g_i) = W(L, \boldsymbol{h}), \quad \boldsymbol{a} = \pi_{\theta}(I_t, l_i, g_i)
\end{equation}
The robot executes $a$ until the current observation $I_t$ visually aligns with the goal $g_i$, at which point the system updates the history $\boldsymbol{h}$ and queries $W$ for the next stage.

In contrast to prevalent VLAs that often discard interaction history to accommodate architectural constraints and hinder compositional reasoning, or video generation frameworks that suffer from the stochasticity of pixel prediction, our formulation adopts a structured hierarchical decomposition. We abstract the continuous task into discrete key milestones $(l_i, g_i)$ rather than relying on the generation of dense future frames. This approach leverages the heuristics of sub-goal invariance and hence avoids ambiguity of video prediction.
Moreover, it provides the policy $\pi_{\theta}$ with explicit references for ``what" to execute and ``how" to verify termination to ensure robustness.

\section{Method}\label{sec:method}

In this section, we describe the two essential modules of \our{}, including the embodied world model and the GoalVLA. For each manipulation task, we divide it into multiple stages, each stage contains an atom skill and a target object or position. Our world model generates the interleaved sequence of text and images as the subtask instruction and goal image for each subtask stage, which we will describe in Sec.~\ref{sec:method_wm}. Next, the GoalVLA would judge the current stage and take the current observation and the subtask and goal images of the current stage as input to predict the current action chunk, this process will be detailed in Sec.~\ref{sec:method_vla}.  Finally, the datasets used for the training of the two modules are illustrated in Sec.~\ref{sec:method_dataset}.

\subsection{World Model Planner}\label{sec:method_wm}

To enable high-level planning, we introduce the Embodied World Model which we denoted as \(\mathcal{W}\), that decomposes the global instruction $L$ into executable milestones $(l_i, g_i)$. By treating visual goals and textual subtasks as a unified multi-modal sequence, \(\mathcal{W}\) autoregressively predicts the next stage of the task conditioned on the generated history.

\paragraph{Model Formulation} We adopt a unified discrete representation by mapping both vision and language into a shared vocabulary $\mathcal{V}$. We utilize the IBQ-Tokenizer \citep{shi2025imagetokenizer} for images and the Qwen3 tokenizer \citep{bai2023qwen} for text, denoted jointly as the tokenization function $\phi$. The multi-view images, which are flattened in a fixed order, and text are converted into a discrete token sequence $S$:
\begin{align}
S &= (\phi(I_0), \phi(L), \phi(l_0), \phi(g_0), \dots, \phi(l_N), \phi(g_N)) \nonumber\\
&= (u_1, u_2, \dots, u_K) \nonumber
\end{align}
where $u_k \in \mathcal{V}$ represents the $k$-th token in the sequence of length $K$. This unifies the global context ($I_0, L$) and the interleaved milestones ($l_i, g_i$) into a single homogeneous input for the transformer.
Based on the unified sequence formulation, the model is trained via standard autoregressive modeling to learn the joint distribution of the task sequence. We optimize the cross-entropy loss $\mathcal{L}$ over the sequence $S$, maximizing the likelihood of the next token given the preceding context:
\begin{equation}
\mathcal{L} = -\sum_{k=1}^{K} \log P(u_k \mid u_{<k}; \theta_{\mathcal{W}})
\end{equation}
where $\theta_{\mathcal{W}}$ denotes the model parameters. Training utilizes teacher forcing with a causal attention mask, ensuring that predictions at step $k$ depend solely on the history $u_{<k}$.

\paragraph{Subtask Planning} During inference, generation proceeds autoregressively via iterative sampling to predict the sequence of milestones. Starting from the initial context $(\phi(I_0), \phi(L))$, the model sequentially predicts tokens for subsequent subtasks and goal images using beam search. Specifically, we maintain a set of $B$ candidate sequences (where $B$ is the beam width). At each step $j$, given the current token history $S_{<j} = (u_1, \ldots, u_{j-1})$, we compute the output logits and expand each candidate with the top-$B$ most probable next tokens. The joint probability of a candidate sequence $S$ is computed as the product of the conditional probabilities of its constituent tokens:
\begin{equation}
P(S) = \prod_{j} P(u_j \mid u_{<j})
\end{equation}
The sampling process repeats until termination tokens are generated for all candidates. We then select the sequence with the highest overall probability and use the inverse tokenizer $\phi^{-1}$ to reconstruct the goal images $g_i$ at the pixel-level. This approach enables the generation of globally coherent milestone sequences conditioned on the initial input, effectively mitigating the risk of suboptimal token-level decisions that can occur with greedy decoding. We implement a continued training on the open-sourced EMU3.5~\citep{cui2025emu3} checkpoint, which is trained on interleaved text-image data, including navigation and manipulation, with 2000 steps. Please see Appendix~\ref{sec:app_impl_detail} for detailed training hyper parameters.

\subsection{Goal Conditioned VLA}\label{sec:method_vla}

Leveraging the planning capability of the World Model, we train a goal images guided Vision-Language-Action (VLA) policy $\pi_\theta$ as a low-level controller for fine-grained manipulation.

\paragraph{Model Formulation} At each time step $t$ in stage $i$, the policy takes the current observation $I_t$, the subtask instruction $l_i$, and the corresponding goal image $g_i$ as input to predict an action chunk $\boldsymbol{a} \in \mathcal{A}^k$ that drives the robot toward the milestone. To fuse visual information effectively, we extend the input representation by concatenating the current observation tokens with the goal image tokens before feeding them into the model backbone.

Following the architecture of $\pi_0$~\citep{black2025pi0}, we train our policy using flow matching objective to generate continuous action trajectories. We define a linear interpolation between a noise sample $\boldsymbol{z} \sim \mathcal{N}(\mathbf{0}, \mathbf{I})$ and the ground-truth action chunk $\boldsymbol{a}$ as $\mathbf{x}_{\tau} = (1 - \tau)\boldsymbol{z} + \tau \boldsymbol{a}$, where $\tau \in [0,1]$ denotes the flow time. The policy predicts the corresponding velocity field $\boldsymbol{v}_{\tau} = \boldsymbol{a} - \boldsymbol{z}$ conditioned on the $(l_i,I_t, g_i)$. The objective is to minimize the expected mean squared error:
\begin{equation}
\mathcal{L}_{\mathrm{FM}} = \mathbb{E}{\tau, \boldsymbol{z}} \left[ \left| \pi_\theta(\mathbf{x}_{\tau}, l_i, I_t, g_i, \tau) - \boldsymbol{v}_{\tau} \right|^2 \right].
\end{equation}

\paragraph{Subtask-Aware Action Padding} Action chunks may straddle subtask boundaries (i.e., include steps that belong to stage $i{+}1$). To prevent the policy from prematurely executing the next stage without an updated goal, we zero-pad the portion of each ground-truth chunk that occurs after the completion of the current milestone, explicitly teaching the policy to stop once the goal $g_i$ is satisfied.

\paragraph{Random Goal Image Offset}

Goal images predicted by the world model may be a few frames early/late relative to the true subtask termination, which can destabilize execution near stage boundaries. To make GoalVLA robust to this misalignment, we define a temporal overlap window around the boundary between subtasks $i$ and $i{+}1$. For training samples inside this window, we randomly use either $g_i$ or $g_{i+1}$ as the goal condition; when $g_{i+1}$ is used, we also relabel the sample as stage $i{+}1$ to avoid zero-padding. For samples in the rest of stage $i$, we randomly sample $g_i$ from the terminal overlap window instead of using a fixed goal frame. This simple augmentation exposes the policy to boundary noise and encourages smooth stage transitions.

\begin{figure}[!t]
    \centering
    \includegraphics[width=\linewidth]{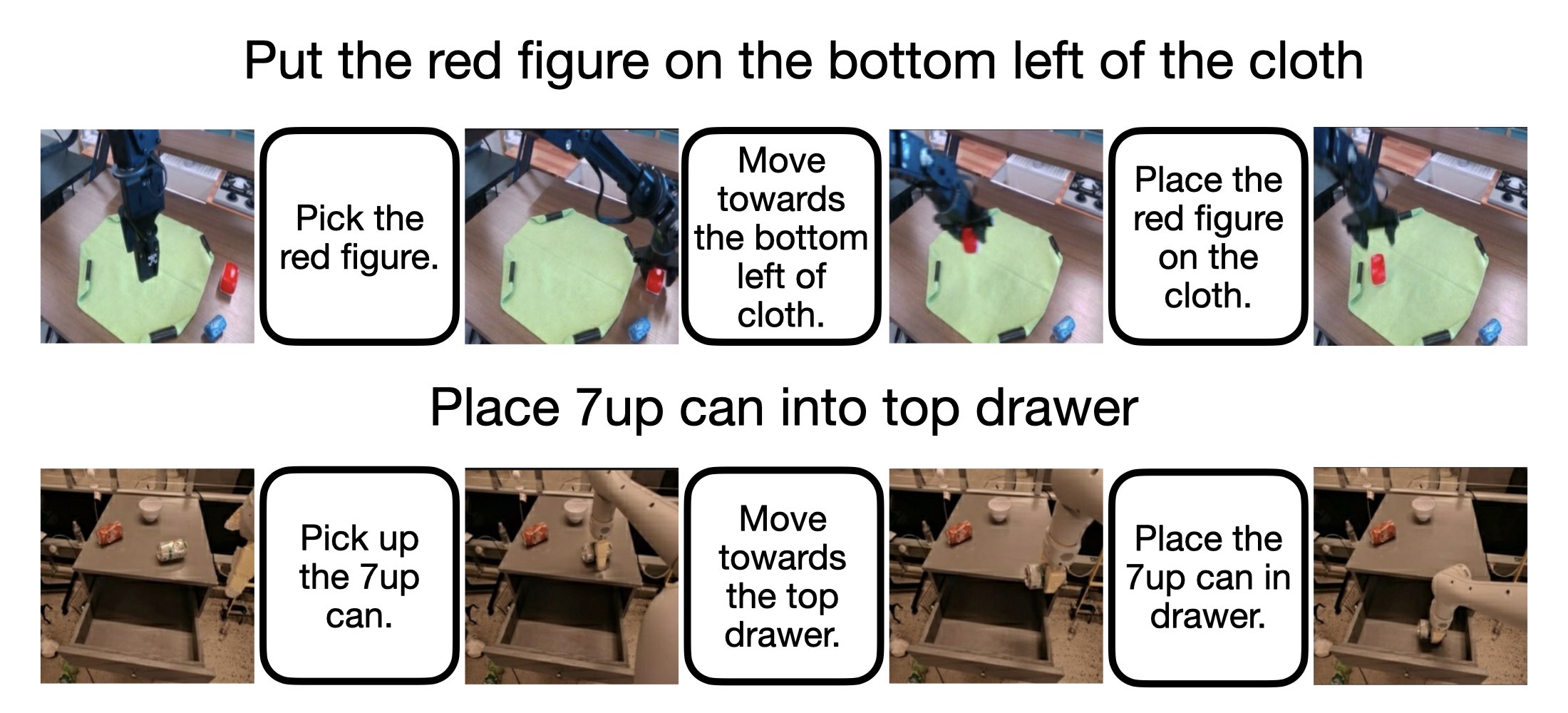}
    \vspace{-6pt}
    \caption{Visualization of samples from our constructed embodied dataset.
    The dataset represents manipulation tasks as interleaved sequences of textual subtask and corresponding visual goals. 
    }
    \vspace{-15pt}
    \label{fig:emu_dataset}
\end{figure}

\subsection{Dataset Curation} \label{sec:method_dataset}

\paragraph{Dataset Source} 
We train the world model on large-scale embodied manipulation data from Open-X-Embodiment~\citep{o'neill2024oxe}, AgiBot World Beta~\citep{bu2025agibot}, and our Mobile Aloha dataset~\citep{fu2024aloha}. Although the corpus exceeds 1M trajectories, annotations include only global instructions and image–action pairs, lacking intermediate milestones $(l_i, g_i)$. We therefore introduce an automated trajectory decomposition pipeline.

\paragraph{Automatic Milestone Labeling} 
The pipeline proceeds in three steps. First, we build a library of 50 atomic skills (e.g., \emph{pick}, \emph{push}) by clustering instruction verbs with Qwen3~\citep{yang2025qwen3}. Second, candidate milestone boundaries are detected via physical state changes, using Ramer–Douglas–Peucker (RDP)~\citep{douglas1973rdp} on motion trajectories and gripper state transitions. Finally, Qwen2.5-VL 72B~\citep{bai2025qwen2p5vl} merges adjacent segments with identical skills and generates natural language subtask descriptions $l_i$.

This process converts 1.2M trajectories into interleaved subtask and goal-image sequences, covering 14 embodiments with multi-view support and totaling 15.2B tokens (Fig.~\ref{fig:emu_dataset}). For AgiBot, we further refine abstract task types into fine-grained instructions.

\paragraph{Any-to-Image Pretraining Data}
We co-train the world model with a 15.0B-token Any-to-Image dataset built from SEED-Data-Edit~\citep{ge2024seed}, WeatherStream~\citep{zhang2023weatherstream}, ShareGPT-4o-Image~\citep{chen2025sharegpt}, etc. See Appendix~\ref{sec:app_dataset_construction} for details.

\section{Experimental Results}\label{sec:exp}

\subsection{Real World Tasks and Setups}
For the training dataset, we collect 2 hours of the pick-place task with five objects: egg, cola can, apple, milk bin, and croissant. 
We selected the \(\pi_0\) model, which features remarkable influence and outstanding performance, as our baseline. To further illustrate the impact of subtask instructions, we modified the training paradigm of \(\pi_0\) by replacing the original instructions in the dataset with subtasks, and we denote this variant as \(\pi_{0}\text{-subtask}\).

\begin{figure}[!t]
    \centering
    \includegraphics[width=1\linewidth]{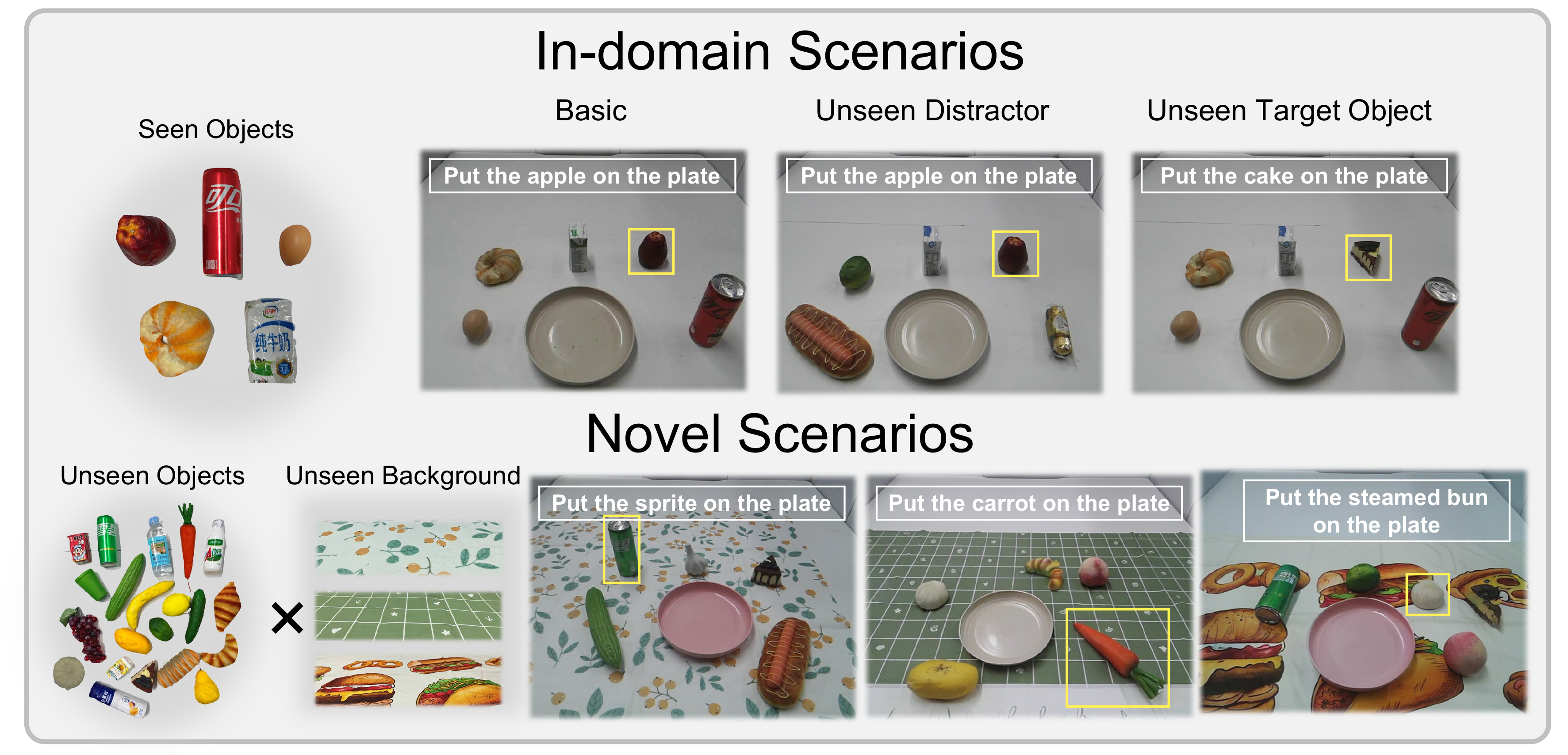}
    \vspace{-6pt}
    \caption{Visualization of real-world experiment setups, including the in-domain and novel scenarios.
    }
    \vspace{-15pt} 
    \label{fig:real_setup_ill}
\end{figure}

As shown in Fig.~\ref{fig:real_setup_ill}, we evaluate the methods over the in-domain and out-of-distribution (ood) scenarios. For in-domain scenarios, despite the basic setup, we further replace the distractors and target objects to unseen objects. With each setting in in-domain scenarios, we have 5 tasks $\times$ 3 scene setup (combination of layout and unseen objects) $=$ 15 scenarios. For ood scenarios, we involve 21 novel objects with different semantic types and sample 5 objects for each scenario, of which one is selected as the target and the other four as the distractors. Then, we have 21 targets $\times$ 3 scene setup (combination of table-cloth and layout) $=$ 63 scenarios. For each scenario, we evaluate the model 3 times and calculate the mean metrics, and thus conduct 78 scenarios (15 in-domain + 63 ood) $\times$ 3 rollouts per scenario $=$ 234 rollouts per policy.

\begin{figure*}[!t]
    \centering
    \includegraphics[width=\textwidth]{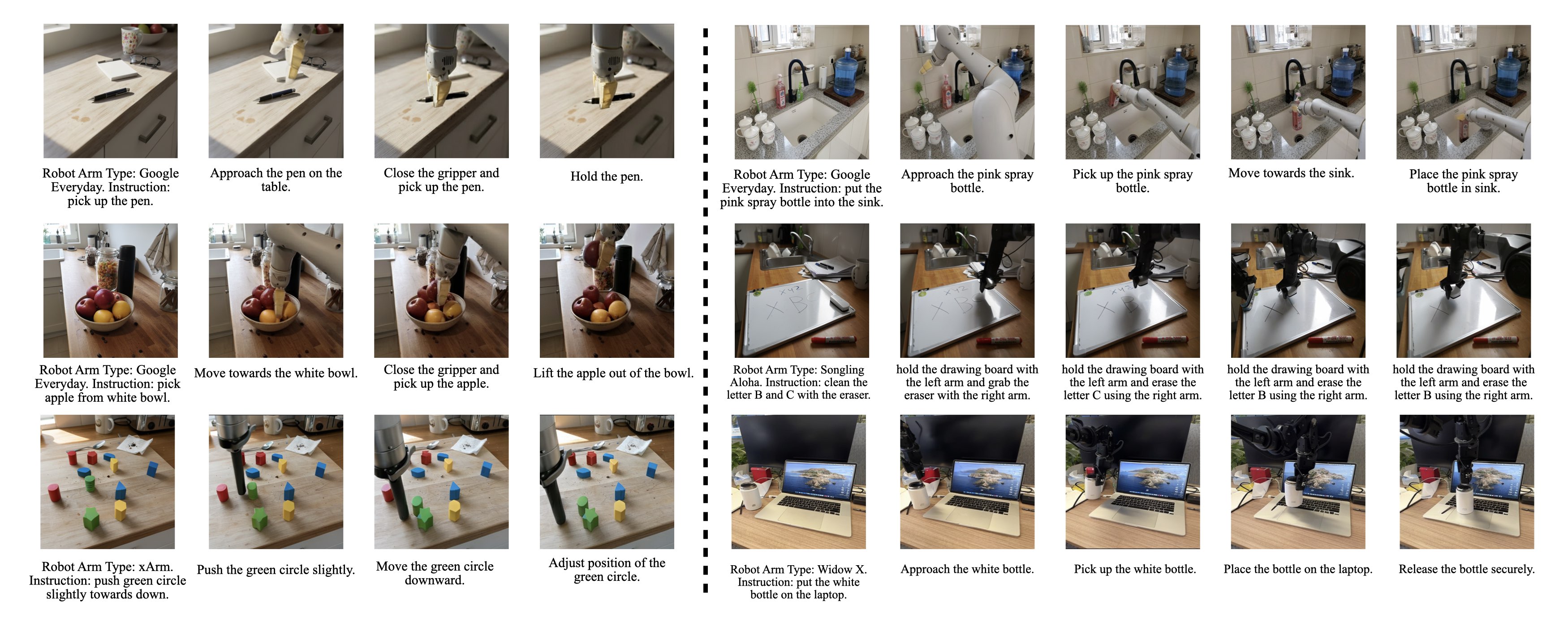}
    \vspace{-6pt}
    \caption{The qualitative results for the subtask and goal image sequences generated by \our{}. 
    See Appendix~\ref{sec:app_vis_unconstrained} for more results.
    }
    \vspace{-10pt}
    \label{fig:bench_res}
\end{figure*}

\begin{figure*}[htbp]
    \centering
    \includegraphics[width=\linewidth]{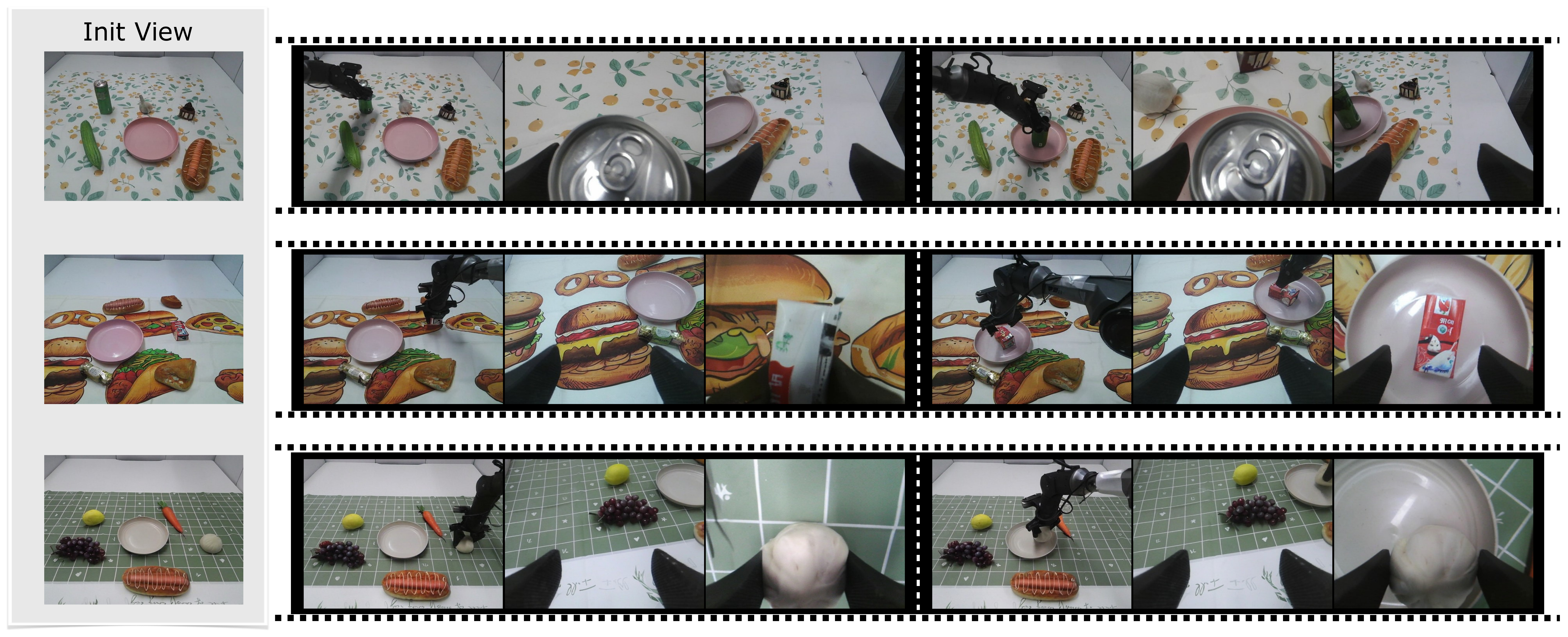}
    \vspace{-6pt}
    \caption{The visualization of the multi-view goal images generated by \our{} in the real robot workspace with novel scenarios (unseen layout, distractor, target object, and background).}
    \vspace{-15pt}
    \label{fig:emu_real_ood} 
\end{figure*}

\subsection{Visual Subgoal–based Task Decomposition}

With the continued training over the above-mentioned large-scale robotics and general dataset, \our{} emerges with the following abilities:
\begin{itemize}

    \item Generate a physically consistent manipulation process over diverse phone captured and synthetic scenes.

    \item Maintain multi-view and physical consistency over diverse backgrounds, objects, and scenarios.

    \item Compose independent skills into entirely unseen long-horizon tasks, while grounding semantic and spatial understanding from language into visual goal images.

    \item Transfer the real robot photos captured by Aloha to other embodiments. 
\end{itemize}

\subsubsection{Qualitative Results on Real Photos Synthetic Scenes}

To evaluate the generalization of \our{}, we generate out-of-distribution initial frames using Nano Banana~\citep{comanici2025gemini2p5}, based on phone-captured images and real robot trajectories. Nano Banana is prompted to preserve object semantic types while varying object appearance, layout, background, and camera viewpoint. As shown in Fig.~\ref{fig:bench_res}, the first image in each panel is an OOD initial frame, which we manually verify and annotate with the corresponding robot embodiment and language instruction.

The subsequent subtask and goal image sequences generated by \our{} demonstrate coherent planning and physically plausible interactions. The model consistently approaches target objects, synthesizes appropriate arm poses, and generates interaction sequences with realistic object motion, shadows, and grasping behavior, rather than artifacts such as object attachment. This indicates that \our{} maintains high-quality generation under significant visual and contextual shifts.

Moreover, \our{} exhibits cross-embodiment generalization. For example, the spray bottle is unseen in RT-1 and the laptop is unseen in Bridge, yet the model successfully transfers object affordances and skill knowledge across embodiments. This suggests that \our{} encodes skills, objects, and embodiments in a shared representation space, enabling linear rather than combinatorial generalization.

Finally, Fig.~\ref{fig:bench_res} (right) shows that when instructed to erase only the letters “B” and “C,” the model correctly identifies and removes the specified regions, despite being trained only on full-board cleaning tasks. This demonstrates that the world model preserves fine-grained world knowledge (e.g., OCR) and generates realistic future states conditioned on precise instructions.

\begin{figure*}[!t]
    \centering
    \includegraphics[width=\textwidth]{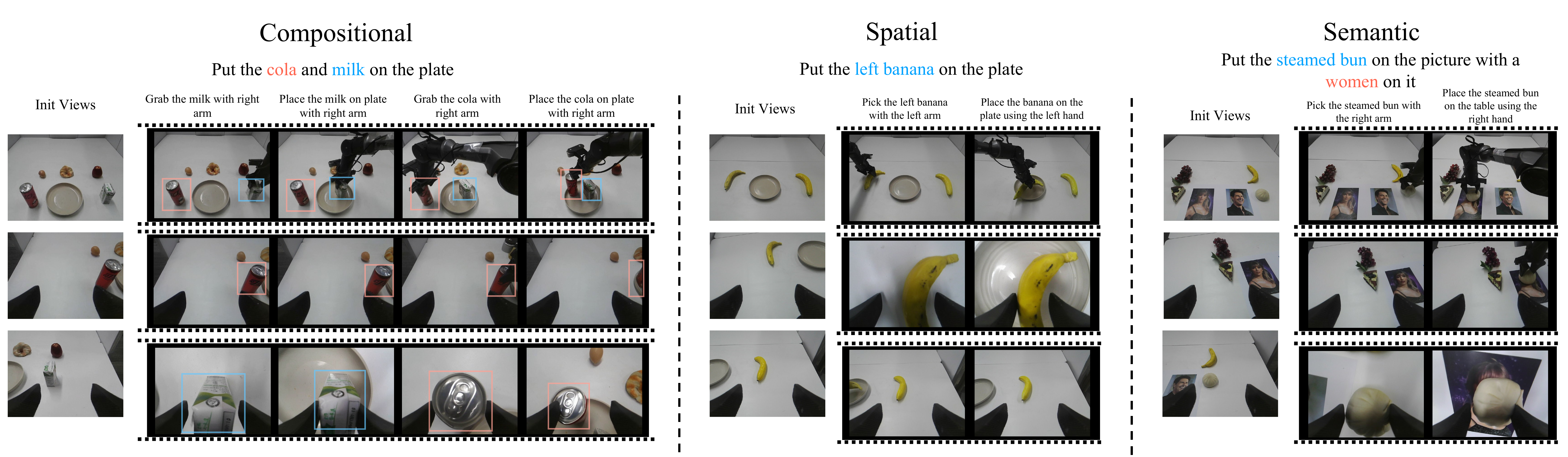}
     \caption{Visualization of \our{} generated sequences over compositional, spatial understanding, and semantic understanding tasks.}
    \vspace{-10pt}
    \label{fig:real_comp}
\end{figure*}

\begin{figure}[!t]
    \centering

    \includegraphics[width=\linewidth]{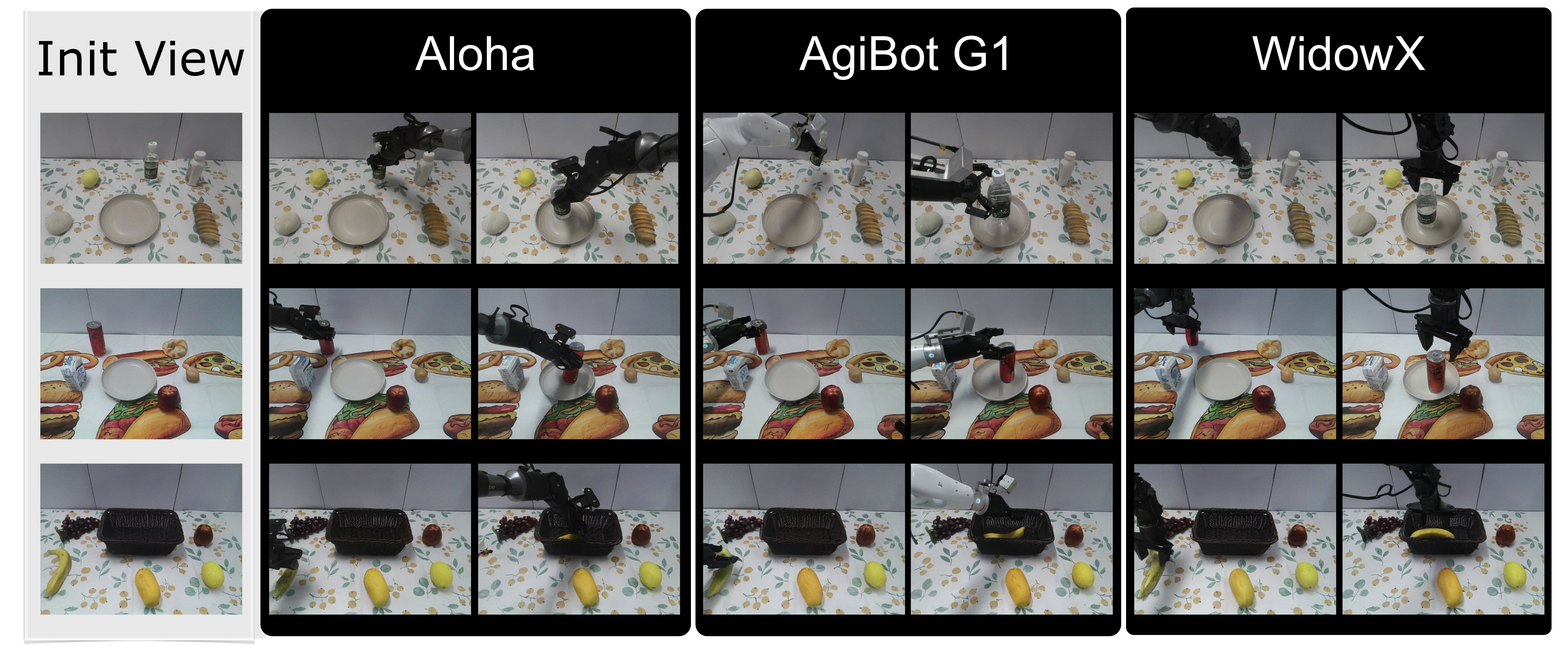}
    \vspace{-6pt}
    \caption{The visualization for cross-embodiment transfer. \our{} could generate high-quality rollouts for different embodiments based on the robot arm type provided in the prompt.
    }
    \vspace{-15pt}
    \label{fig:real_cross}
\end{figure}
\subsubsection{Qualitative Results on Real-Robot Scenarios}

As shown in Fig.~\ref{fig:emu_real_ood}, \our{} exhibits strong multi-view generation, producing spatially consistent goal images across head, left-wrist, and right-wrist cameras. Even under unseen scene layouts, objects, backgrounds, and viewpoints, the generated sequences maintain coherent spatial relationships between target objects and surrounding distractors across all views, including top-down perspectives. This indicates robust 3D spatial reasoning and generalization to novel multi-view observations.

The interleaved goal sequences accurately reflect the temporal structure of manipulation tasks such as pick-and-place. Across views, the generated goals depict consistent arm and gripper configurations, including approach trajectories, grasp poses, and object relocation, ensuring physically plausible and actionable goal states.

Multi-view visual goals provide complementary guidance beyond textual subtasks. While language specifies semantic intent, goal images convey precise spatial constraints, such as relative positions and orientations, which are difficult to express textually. This combined representation enables GoalVLA to better interpret task requirements and improves robustness in downstream control.

Despite being trained on single-arm, single-object tasks in Mobile Aloha, \our{} generalizes to multi-object and dual-arm compositions. As shown in Fig.~\ref{fig:real_comp}, the model generates coherent multi-step plans under different arm usage constraints, with consistent object motion observed across all camera views, even when such cross-arm observations are absent from the training data. This further demonstrates strong multi-view consistency.

Finally, \our{} transfers across embodiments. When prompted with Mobile Aloha head-camera images and different robot arms, the model successfully generates interleaved goal sequences for Aloha, AgiBot G1, and WidowX in unseen environments (Fig.~\ref{fig:real_cross}). We additionally evaluate more complex tasks, such as T-shirt folding and object packing, with results provided in Appendix~\ref{sec:app_vis_unconstrained}.

\begin{table*}[!t]
\tiny
\caption{Comparison with baseline methods over the basic, unseen distractors and unseen target object setting for the collected 5 training tasks. The \textit{App} denotes the approach success rate, and \textit{Suc} denotes the execution success rate.}
\centering
\renewcommand{\arraystretch}{1.1}
\resizebox{\textwidth}{!}{
\begin{tabular}{cccccccccccccc}
\hline
\multirow{2}{*}{Setup} & \multirow{2}{*}{Method} & \multicolumn{2}{c}{Task 1 (cola can)} & \multicolumn{2}{c}{Task 2 (egg)} & \multicolumn{2}{c}{Task 3 (bread)} & \multicolumn{2}{c}{Task 4 (apple)} & \multicolumn{2}{c}{Task 5 (milk)} & \multicolumn{2}{c}{Avg} \\ 
 &  & App & Suc & App & Suc & App & Suc & App & Suc & App & Suc & App & Suc \\ \hline
\multicolumn{1}{c}{\multirow{3}{*}{Basic}} & $\pi_{\scalebox{0.8}{0}}$ & 1.0 & 1.0 & 1.0 & 1.0 & 1.0 & 1.0 & 1.0 & 0.89 & 1.0 & 0.89 & 1.0 & \textbf{0.96} \\ 
\multicolumn{1}{c}{} & $\pi_{\scalebox{0.8}{0}}$-subtask & 1.0 & 0.89 & 1.0 & 1.0 & 1.0 & 1.0 & 1.0 & 0.67 & 1.0 & 1.0 & 1.0 & 0.91 \\ 
\multicolumn{1}{c}{} & Ours & 1.0 & 0.89 & 1.0 & 1.0 & 1.0 & 1.0 & 1.0 & 0.78 & 1.0 & 1.0 & 1.0 & 0.93 \\ \hline
\multicolumn{1}{c}{\multirow{3}{*}{Unseen Distractor}} & $\pi_{\scalebox{0.8}{0}}$ & 1.0 & 1.0 & 1.0 & 1.0 & 1.0 & 0.56 & 1.0 & 0.11 & 1.0 & 1.0 & 1.0 & 0.73 \\ 
\multicolumn{1}{c}{} & $\pi_{\scalebox{0.8}{0}}$-subtask & 1.0 & 1.0 & 1.0 & 0.78 & 1.0 & 1.0 & 1.0 & 0.33 & 1.0 & 0.78 & 1.0 & 0.78 \\ 
\multicolumn{1}{c}{} & Ours & 1.0 & 1.0 & 1.0 & 0.56 & 1.0 & 1.0 & 1.0 & 0.56 & 1.0 & 1.0 & 1.0 & \textbf{0.82} \\ \hline
\multicolumn{1}{c}{\multirow{3}{*}{Unseen Target}} & $\pi_{\scalebox{0.8}{0}}$ & 0.33 & 0.0 & 0.67 & 0.0 & 0.33 & 0.0 & 0.33 & 0.22 & 0.33 & 0.0 & 0.40 & 0.04 \\ 
\multicolumn{1}{c}{} & $\pi_{\scalebox{0.8}{0}}$-subtask & 0.33 & 0.33 & 1.0 & 0.56 & 1.0 & 0.33 & 0.33 & 0.0 & 1.0 & 0.33 & 0.73 & 0.31 \\ 
\multicolumn{1}{c}{} & Ours & 1.0 & 0.67 & 1.0 & 1.0 & 1.0 & 0.67 & 1.0 & 0.67 & 1.0 & 0.33 & 1.0 & \textbf{0.67} \\ \hline
\end{tabular}
\label{tab:basic}
\vspace{-15pt}
}
\end{table*}
\subsection{Goal Conditioned VLA}

\subsubsection{Quantitative Analysis}
To evaluate robustness and generalization, we consider two settings: (i) replacing training-case distractors with unseen distractors, and (ii) substituting target objects with unseen targets while keeping the scene layout and instruction structure unchanged. Quantitative results are reported in Table~\ref{tab:basic}. In unseen-distractor settings, all models experience some degradation, particularly for tasks requiring precise grasping (e.g., eggs and apples), but our model consistently achieves the highest overall success rate, demonstrating superior robustness.

\begin{figure*}[!t]
    \centering
    \includegraphics[width=\linewidth]{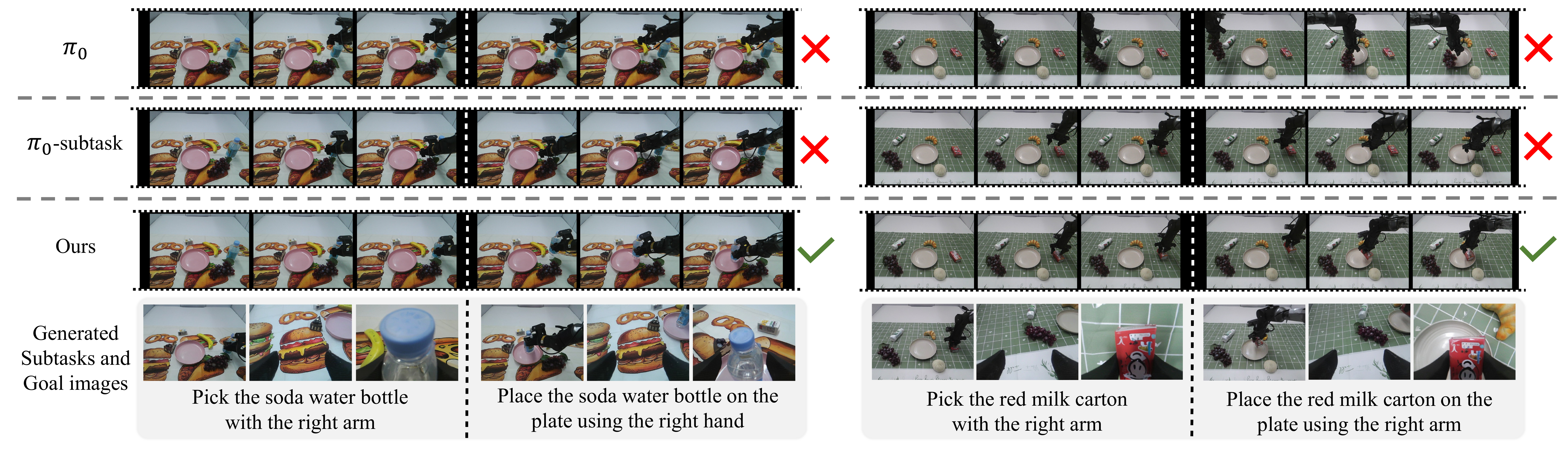}
    \vspace{-6pt}
    \caption{The comparison of execution visualization results in novel scenarios.
    }
    \vspace{-15pt}
    \label{fig:case_comparison_compact}
\end{figure*}

In unseen-target settings, the policy $\pi_0$ performs poorly in both approach and execution, highlighting the difficulty of grounding novel object names in vision. $\pi_0$-subtask improves approach success via subtask guidance but still fails during execution, revealing the limitations of language-only conditioning. In contrast, our model achieves a 100\% approach success rate and significantly higher execution success, enabled by goal-image guidance that provides explicit spatial cues for accurate grasping and execution on unseen objects.

\begin{figure}[!t]
    \centering
    \includegraphics[width=\linewidth]{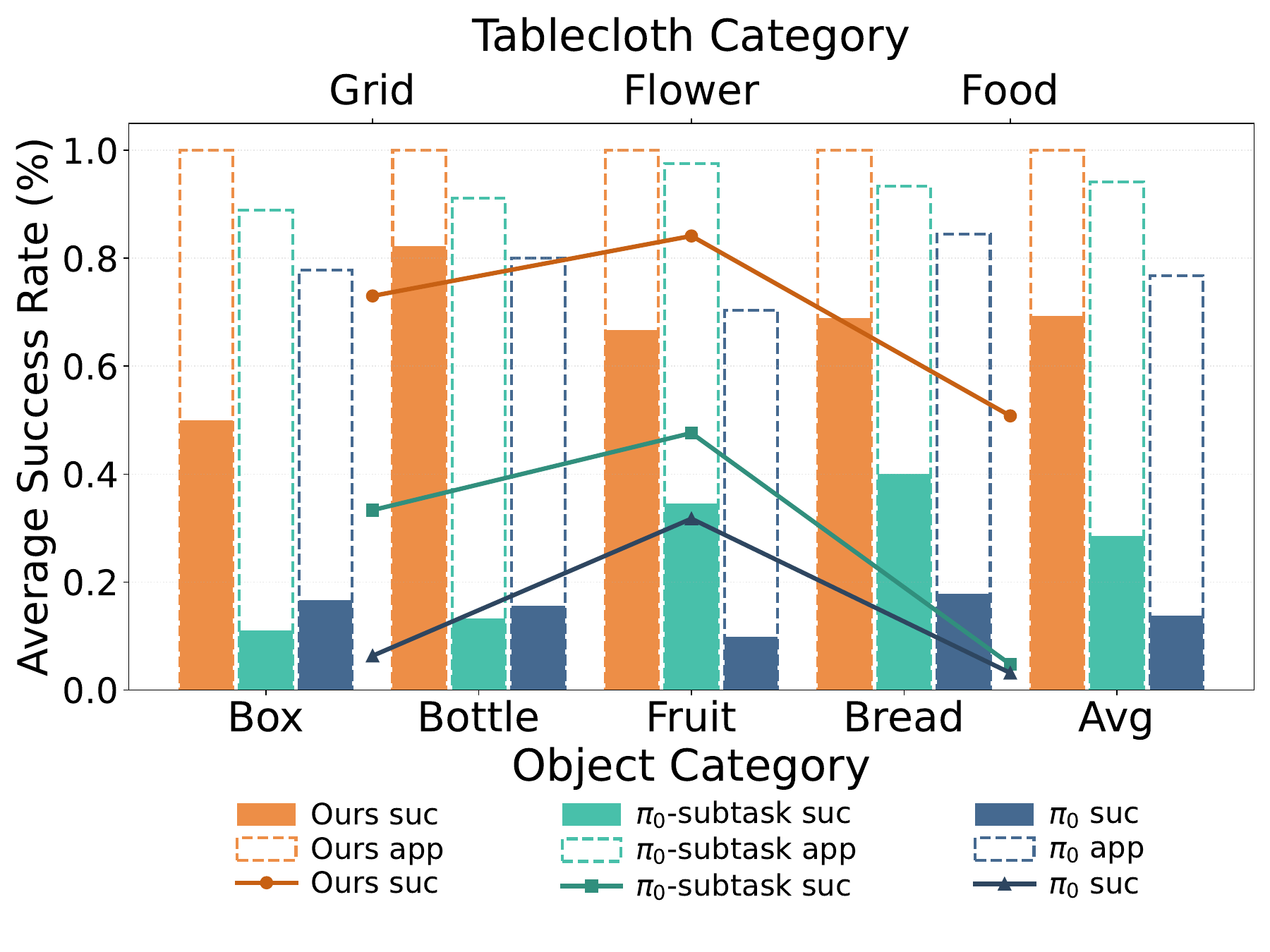}
    \vspace{-6pt}
    \caption{Comparison with baseline methods on unseen scenarios over different target object categories and backgrounds. The bar chart corresponds to the bottom x-axis, and the line chart to the top x-axis.
    }
    \vspace{-15pt}
    \label{fig:mixture_res}
\end{figure}

We further investigate the generalization limits by employing highly challenging scenarios with patterned tablecloths and entirely unseen objects across multiple categories, including fruits, breads, boxes, and bottles. The evaluation results are shown in Fig.~\ref{fig:mixture_res}. Despite substantial visual interference, our model maintains a high success rate, whereas baselines fail to execute accurate grasps despite reasonable approach performance. Background-specific analysis shows that semantically rich patterns cause performance drops for all models; however, our model exhibits the smallest degradation, confirming its stronger robustness to complex visual backgrounds.

\subsubsection{Qualitative Analysis}

As shown in Fig.~\ref{fig:case_comparison_compact}, we qualitatively compare three policies in novel scenarios with unseen objects, table layouts, and backgrounds. In these settings, the SOTA VLA $\pi_0$ and its subtask-guided variant \(\pi_{0}\text{-subtask}\) fail, while our method succeeds. In the first scenario on the left of Fig.~\ref{fig:case_comparison_compact}, all policies initially approach the soda bottle; however, without visual goal guidance, both $\pi_0$ and \(\pi_{0}\text{-subtask}\) baseline fail to grasp it. This reflects a common limitation of learning-based VLAs: when the target object is unseen, the policy interpolates action chunks from visually similar training objects. With only two hours of real-robot data, the closest object is a cola can, causing the baselines to adopt a horizontal can-grasping posture rather than a vertical bottle grasp, leading to failure under distribution shift.

In contrast, our method generates temporally aligned, multi-view goal images that explicitly encode the interaction region and desired end-effector pose. Conditioned on these goal images, GoalVLA accurately adjusts both position and posture, successfully grasping the unseen bottle and completing the task. This highlights the effectiveness of our world-model-driven hierarchy, where the world model supplies precise visual interaction cues and GoalVLA follows combined textual and visual guidance to produce robust action chunks.

A similar pattern appears in the second scenario on the right of Fig.~\ref{fig:case_comparison_compact}. When instructed to grasp the red milk carton, $\pi_0$ misidentifies the target and grasps a grape. Refining the instruction via subtask decomposition improves target localization but still fails to execute a precise grasp, even with subtask guidance. In contrast, our method uses goal-image spatial cues to guide arm selection, end-effector alignment, and grasp execution, enabling successful manipulation of the unseen object.

Overall, these results demonstrate that multi-view visual goals deliver substantially richer spatial guidance compared to language alone. By conditioning low-level control on goal images, our framework achieves superior robustness to unseen objects, novel semantic information, and visual interference.  Additionally, as shown in Fig.~\ref{fig:real_comp}, \our{} generates coherent textual and visual guidance for compositional tasks, demonstrating understanding of object attributes and spatial relations. It also correctly interprets instructions that incorporate spatial relationships and semantic meaning. Additional results and real-world executions are provided in Appendix~\ref{sec:app_emerging_capability} and Appendix~\ref{sec:app_execution_result}.

\section{Conclusion}\label{sec:conclusion}
We propose a generative hierarchical framework that unifies world modeling and visual goal–conditioned policy learning for long-horizon robotic manipulation. By synthesizing textual subtasks and visual subgoals, our method effectively bridges high-level semantic reasoning and low-level control through an explicit, verifiable intermediate representation, providing spatial guidance beyond what language alone can offer.

Real-world experiments demonstrate significant gains in data efficiency, robustness, and out-of-distribution generalization. The learned world model produces physically plausible multi-view goal images across unseen objects and scenes, while the goal-conditioned VLA reliably follows these visual subgoals to perform precise manipulation. These results highlight visual goal synthesis as a powerful decomposition for scalable and generalizable robot learning, enabling tighter integration between world models and embodied control.

\section{Limitations}\label{sec:limit}
While \our{} shows strong potential for cross-embodiment generalization and complex tasks (e.g., liquid pouring and deformable object folding), our real-world evaluation is currently limited to pick-and-place tasks. Extending validation to more diverse and longer-horizon manipulation scenarios remains an important direction for future work. Additionally, the generated visual subgoals may exhibit minor spatial deviations from the true subtask endpoints; although infrequent, larger deviations can lead to execution failures in GoalVLA. Finally, while visual goals transfer useful interaction priors across embodiments, successful execution still depends on the presence of corresponding interaction patterns in the real-robot training data. Increasing the diversity of skills and tasks during GoalVLA fine-tuning is therefore essential to further improve robustness and generalization.

\clearpage
\bibliographystyle{plainnat}
\bibliography{references}

@article{ge2024seed,
  title={Seed-x: Multimodal models with unified multi-granularity comprehension and generation},
  author={Ge, Yuying and Zhao, Sijie and Zhu, Jinguo and Ge, Yixiao and Yi, Kun and Song, Lin and Li, Chen and Ding, Xiaohan and Shan, Ying},
  journal={arXiv preprint arXiv:2404.14396},
  year={2024}
}

@article{chen2025sharegpt,
  title={ShareGPT-4o-Image: Aligning Multimodal Models with GPT-4o-Level Image Generation},
  author={Chen, Junying and Cai, Zhenyang and Chen, Pengcheng and Chen, Shunian and Ji, Ke and Wang, Xidong and Yang, Yunjin and Wang, Benyou},
  journal={arXiv preprint arXiv:2506.18095},
  year={2025}
}

@inproceedings{xiao2025omnigen,
  title={Omnigen: Unified image generation},
  author={Xiao, Shitao and Wang, Yueze and Zhou, Junjie and Yuan, Huaying and Xing, Xingrun and Yan, Ruiran and Li, Chaofan and Wang, Shuting and Huang, Tiejun and Liu, Zheng},
  booktitle={Proceedings of the Computer Vision and Pattern Recognition Conference},
  pages={13294--13304},
  year={2025}
}

@article{yu2024promptfix,
  title={Promptfix: You prompt and we fix the photo},
  author={Yu, Yongsheng and Zeng, Ziyun and Hua, Hang and Fu, Jianlong and Luo, Jiebo},
  journal={arXiv preprint arXiv:2405.16785},
  year={2024}
}

@inproceedings{zhang2023weatherstream,
  title={Weatherstream: Light transport automation of single image deweathering},
  author={Zhang, Howard and Ba, Yunhao and Yang, Ethan and Mehra, Varan and Gella, Blake and Suzuki, Akira and Pfahnl, Arnold and Chandrappa, Chethan Chinder and Wong, Alex and Kadambi, Achuta},
  booktitle={Proceedings of the IEEE/CVF conference on computer vision and pattern recognition},
  pages={13499--13509},
  year={2023}
}

@article{ye2025imgedit,
  title={Imgedit: A unified image editing dataset and benchmark},
  author={Ye, Yang and He, Xianyi and Li, Zongjian and Lin, Bin and Yuan, Shenghai and Yan, Zhiyuan and Hou, Bohan and Yuan, Li},
  journal={arXiv preprint arXiv:2505.20275},
  year={2025}
}

@article{wang2025gpt,
  title={Gpt-image-edit-1.5 m: A million-scale, gpt-generated image dataset},
  author={Wang, Yuhan and Yang, Siwei and Zhao, Bingchen and Zhang, Letian and Liu, Qing and Zhou, Yuyin and Xie, Cihang},
  journal={arXiv preprint arXiv:2507.21033},
  year={2025}
}

@article{wu2025omnigen2,
  title={OmniGen2: Exploration to Advanced Multimodal Generation},
  author={Wu, Chenyuan and Zheng, Pengfei and Yan, Ruiran and Xiao, Shitao and Luo, Xin and Wang, Yueze and Li, Wanli and Jiang, Xiyan and Liu, Yexin and Zhou, Junjie and others},
  journal={arXiv preprint arXiv:2506.18871},
  year={2025}
}

@article{Chen2025MultiRefCI,
  title={MultiRef: Controllable Image Generation with Multiple Visual References},
  author={Ruoxi Chen and Dongping Chen and Siyuan Wu and Sinan Wang and Shiyun Lang and Petr Sushko and Gaoyang Jiang and Yao Wan and Ranjay Krishna},
  journal={ArXiv},
  year={2025},
  volume={abs/2508.06905},
  url={https://api.semanticscholar.org/CorpusID:280565765}
}

@inproceedings{jang2022bc,
  title={Bc-z: Zero-shot task generalization with robotic imitation learning},
  author={Jang, Eric and Irpan, Alex and Khansari, Mohi and Kappler, Daniel and Ebert, Frederik and Lynch, Corey and Levine, Sergey and Finn, Chelsea},
  booktitle={Conference on Robot Learning},
  pages={991--1002},
  year={2022},
  organization={PMLR}
}

@inproceedings{shridhar2022cliport,
  title={Cliport: What and where pathways for robotic manipulation},
  author={Shridhar, Mohit and Manuelli, Lucas and Fox, Dieter},
  booktitle={Conference on robot learning},
  pages={894--906},
  year={2022},
  organization={PMLR}
}

@inproceedings{brohan2023rt1, 
    AUTHOR    = {Anthony Brohan AND Noah Brown AND Justice Carbajal AND Yevgen Chebotar AND Joseph Dabis AND Chelsea Finn AND Keerthana Gopalakrishnan AND Karol Hausman AND Alexander Herzog AND Jasmine Hsu AND Julian Ibarz AND Brian Ichter AND Alex Irpan AND Tomas Jackson AND Sally Jesmonth AND Nikhil Joshi AND Ryan Julian AND Dmitry Kalashnikov AND Yuheng Kuang AND Isabel Leal AND Kuang-Huei Lee AND Sergey Levine AND Yao Lu AND Utsav Malla AND Deeksha Manjunath AND Igor Mordatch AND Ofir Nachum AND Carolina Parada AND Jodilyn  Peralta AND Emily Perez AND Karl Pertsch AND Jornell  Quiambao AND Kanishka Rao AND Michael S Ryoo AND Grecia  Salazar AND Pannag R Sanketi AND Kevin  Sayed AND Jaspiar  Singh AND Sumedh  Sontakke AND Austin  Stone AND Clayton  Tan AND Huong  Tran AND Vincent Vanhoucke AND Steve  Vega AND Quan H Vuong AND Fei Xia AND Ted Xiao AND Peng Xu AND Sichun Xu AND Tianhe Yu AND Brianna  Zitkovich}, 
    TITLE     = {{RT-1: Robotics Transformer for Real-World Control at Scale}}, 
    BOOKTITLE = {Proceedings of Robotics: Science and Systems}, 
    YEAR      = {2023}, 
    ADDRESS   = {Daegu, Republic of Korea}, 
    MONTH     = {July}, 
    DOI       = {10.15607/RSS.2023.XIX.025} 
}

@inproceedings{zitkovich2023rt2,
  title={Rt-2: Vision-language-action models transfer web knowledge to robotic control},
  author={Zitkovich, Brianna and Yu, Tianhe and Xu, Sichun and Xu, Peng and Xiao, Ted and Xia, Fei and Wu, Jialin and Wohlhart, Paul and Welker, Stefan and Wahid, Ayzaan and others},
  booktitle={Conference on Robot Learning},
  pages={2165--2183},
  year={2023},
  organization={PMLR}
}

@INPROCEEDINGS{black2025pi0, 
    AUTHOR    = {Kevin Black AND Noah Brown AND Danny Driess AND Adnan Esmail AND Michael Robert Equi AND Chelsea Finn AND Niccolo Fusai AND Lachy Groom AND Karol Hausman AND Brian Ichter AND Szymon Jakubczak AND Tim Jones AND Liyiming Ke AND Sergey Levine AND Adrian Li-Bell AND Mohith Mothukuri AND Suraj Nair AND Karl Pertsch AND Lucy Xiaoyang Shi AND Laura Smith AND James Tanner AND Quan Vuong AND Anna Walling AND Haohuan Wang AND Ury Zhilinsky}, 
    TITLE     = {{$\pi_0$: A Vision-Language-Action Flow Model for General Robot Control}}, 
    BOOKTITLE = {Proceedings of Robotics: Science and Systems}, 
    YEAR      = {2025}, 
    ADDRESS   = {LosAngeles, CA, USA}, 
    MONTH     = {June}, 
    DOI       = {10.15607/RSS.2025.XXI.010} 
}

@article{figure2024helix,
  title={Helix: A vision-language-action model for generalist humanoid control},
  author={Figure, AI},
  journal={Figure AI News},
  year={2024}
}

@article{jiang2025galaxea,
  title={Galaxea open-world dataset and g0 dual-system vla model},
  author={Jiang, Tao and Yuan, Tianyuan and Liu, Yicheng and Lu, Chenhao and Cui, Jianning and Liu, Xiao and Cheng, Shuiqi and Gao, Jiyang and Xu, Huazhe and Zhao, Hang},
  journal={arXiv preprint arXiv:2509.00576},
  year={2025}
}

@inproceedings{wu2024gr1,
  author       = {Hongtao Wu and
                  Ya Jing and
                  Chilam Cheang and
                  Guangzeng Chen and
                  Jiafeng Xu and
                  Xinghang Li and
                  Minghuan Liu and
                  Hang Li and
                  Tao Kong},
  title        = {Unleashing Large-Scale Video Generative Pre-training for Visual Robot
                  Manipulation},
  booktitle    = {The Twelfth International Conference on Learning Representations,
                  {ICLR} 2024, Vienna, Austria, May 7-11, 2024},
  year         = {2024},
  url          = {https://openreview.net/forum?id=NxoFmGgWC9},
  bibsource    = {dblp computer science bibliography, https://dblp.org}
}

@article{du2023learning,
  title={Learning universal policies via text-guided video generation},
  author={Du, Yilun and Yang, Sherry and Dai, Bo and Dai, Hanjun and Nachum, Ofir and Tenenbaum, Josh and Schuurmans, Dale and Abbeel, Pieter},
  journal={Advances in neural information processing systems},
  volume={36},
  pages={9156--9172},
  year={2023}
}

@article{escontrela2023viper,
  title={Video prediction models as rewards for reinforcement learning},
  author={Escontrela, Alejandro and Adeniji, Ademi and Yan, Wilson and Jain, Ajay and Peng, Xue Bin and Goldberg, Ken and Lee, Youngwoon and Hafner, Danijar and Abbeel, Pieter},
  journal={Advances in Neural Information Processing Systems},
  volume={36},
  pages={68760--68783},
  year={2023}
}

@article{chen2025tevir,
  title={TeViR: Text-to-Video Reward with Diffusion Models for Efficient Reinforcement Learning},
  author={Chen, Yuhui and Li, Haoran and Jiang, Zhennan and Wen, Haowei and Zhao, Dongbin},
  journal={arXiv preprint arXiv:2505.19769},
  year={2025}
}

@article{hafner2025dreamer,
  title={Mastering diverse control tasks through world models},
  author={Hafner, Danijar and Pasukonis, Jurgis and Ba, Jimmy and Lillicrap, Timothy},
  journal={Nature},
  pages={1--7},
  year={2025},
  publisher={Nature Publishing Group UK London}
}

@InProceedings{bruce2024genie,
  title = 	 {Genie: Generative Interactive Environments},
  author =       {Bruce, Jake and Dennis, Michael D and Edwards, Ashley and Parker-Holder, Jack and Shi, Yuge and Hughes, Edward and Lai, Matthew and Mavalankar, Aditi and Steigerwald, Richie and Apps, Chris and Aytar, Yusuf and Bechtle, Sarah Maria Elisabeth and Behbahani, Feryal and Chan, Stephanie C.Y. and Heess, Nicolas and Gonzalez, Lucy and Osindero, Simon and Ozair, Sherjil and Reed, Scott and Zhang, Jingwei and Zolna, Konrad and Clune, Jeff and Freitas, Nando De and Singh, Satinder and Rockt\"{a}schel, Tim},
  booktitle = 	 {Proceedings of the 41st International Conference on Machine Learning},
  pages = 	 {4603--4623},
  year = 	 {2024},
  editor = 	 {Salakhutdinov, Ruslan and Kolter, Zico and Heller, Katherine and Weller, Adrian and Oliver, Nuria and Scarlett, Jonathan and Berkenkamp, Felix},
  volume = 	 {235},
  series = 	 {Proceedings of Machine Learning Research},
  month = 	 {21--27 Jul},
  publisher =    {PMLR},
  url = 	 {https://proceedings.mlr.press/v235/bruce24a.html}
}

@inproceedings{ye2025lapa,
  author       = {Seonghyeon Ye and
                  Joel Jang and
                  Byeongguk Jeon and
                  Se June Joo and
                  Jianwei Yang and
                  Baolin Peng and
                  Ajay Mandlekar and
                  Reuben Tan and
                  Yu{-}Wei Chao and
                  Bill Yuchen Lin and
                  Lars Liden and
                  Kimin Lee and
                  Jianfeng Gao and
                  Luke Zettlemoyer and
                  Dieter Fox and
                  Minjoon Seo},
  title        = {Latent Action Pretraining from Videos},
  booktitle    = {The Thirteenth International Conference on Learning Representations,
                  {ICLR} 2025, Singapore, April 24-28, 2025},
  year         = {2025},
  url          = {https://openreview.net/forum?id=VYOe2eBQeh},
  bibsource    = {dblp computer science bibliography, https://dblp.org}
}

@INPROCEEDINGS{bu2025univla, 
    AUTHOR    = {Qingwen Bu AND Yanting Yang AND Jisong Cai AND Shenyuan Gao AND Guanghui Ren AND Maoqing Yao AND Ping Luo AND Hongyang Li}, 
    TITLE     = {{Learning to Act Anywhere with Task-centric Latent Actions}}, 
    BOOKTITLE = {Proceedings of Robotics: Science and Systems}, 
    YEAR      = {2025}, 
    ADDRESS   = {LosAngeles, CA, USA}, 
    MONTH     = {June}, 
    DOI       = {10.15607/RSS.2025.XXI.014} 
}

@inproceedings{liang2023codeaspolicy,
  author       = {Jacky Liang and
                  Wenlong Huang and
                  Fei Xia and
                  Peng Xu and
                  Karol Hausman and
                  Brian Ichter and
                  Pete Florence and
                  Andy Zeng},
  title        = {Code as Policies: Language Model Programs for Embodied Control},
  booktitle    = {{IEEE} International Conference on Robotics and Automation, {ICRA}
                  2023, London, UK, May 29 - June 2, 2023},
  pages        = {9493--9500},
  publisher    = {{IEEE}},
  year         = {2023},
  url          = {https://doi.org/10.1109/ICRA48891.2023.10160591},
  doi          = {10.1109/ICRA48891.2023.10160591},
  bibsource    = {dblp computer science bibliography, https://dblp.org}
}

@inproceedings{singh2023progprompt,
  author       = {Ishika Singh and
                  Valts Blukis and
                  Arsalan Mousavian and
                  Ankit Goyal and
                  Danfei Xu and
                  Jonathan Tremblay and
                  Dieter Fox and
                  Jesse Thomason and
                  Animesh Garg},
  title        = {ProgPrompt: Generating Situated Robot Task Plans using Large Language
                  Models},
  booktitle    = {{IEEE} International Conference on Robotics and Automation, {ICRA}
                  2023, London, UK, May 29 - June 2, 2023},
  pages        = {11523--11530},
  publisher    = {{IEEE}},
  year         = {2023},
  url          = {https://doi.org/10.1109/ICRA48891.2023.10161317},
  doi          = {10.1109/ICRA48891.2023.10161317},
  bibsource    = {dblp computer science bibliography, https://dblp.org}
}

@article{liu2023llmp,
  title={Llm+ p: Empowering large language models with optimal planning proficiency},
  author={Liu, Bo and Jiang, Yuqian and Zhang, Xiaohan and Liu, Qiang and Zhang, Shiqi and Biswas, Joydeep and Stone, Peter},
  journal={arXiv preprint arXiv:2304.11477},
  year={2023}
}

@inproceedings{rana2023sayplan,
  author       = {Krishan Rana and
                  Jesse Haviland and
                  Sourav Garg and
                  Jad Abou{-}Chakra and
                  Ian D. Reid and
                  Niko S{\"{u}}nderhauf},
  editor       = {Jie Tan and
                  Marc Toussaint and
                  Kourosh Darvish},
  title        = {SayPlan: Grounding Large Language Models using 3D Scene Graphs for
                  Scalable Robot Task Planning},
  booktitle    = {Conference on Robot Learning, CoRL 2023, 6-9 November 2023, Atlanta,
                  GA, {USA}},
  series       = {Proceedings of Machine Learning Research},
  volume       = {229},
  pages        = {23--72},
  publisher    = {{PMLR}},
  year         = {2023},
  url          = {https://proceedings.mlr.press/v229/rana23a.html},
  bibsource    = {dblp computer science bibliography, https://dblp.org}
}

@article{nasiriany2019leap,
  title={Planning with goal-conditioned policies},
  author={Nasiriany, Soroush and Pong, Vitchyr and Lin, Steven and Levine, Sergey},
  journal={Advances in neural information processing systems},
  volume={32},
  year={2019}
}

@article{ye2025vla,
  title={Vla-r1: Enhancing reasoning in vision-language-action models},
  author={Ye, Angen and Zhang, Zeyu and Wang, Boyuan and Wang, Xiaofeng and Zhang, Dapeng and Zhu, Zheng},
  journal={arXiv preprint arXiv:2510.01623},
  year={2025}
}

@article{intelligence2025pi_,
  title={$pi_{0.5}$: a Vision-Language-Action Model with Open-World Generalization},
  author={Intelligence, Physical and Black, Kevin and Brown, Noah and Darpinian, James and Dhabalia, Karan and Driess, Danny and Esmail, Adnan and Equi, Michael and Finn, Chelsea and Fusai, Niccolo and others},
  journal={arXiv preprint arXiv:2504.16054},
  year={2025}
}

@article{yang2025roboenvision,
  title={RoboEnvision: A Long-Horizon Video Generation Model for Multi-Task Robot Manipulation},
  author={Yang, Liudi and Bai, Yang and Eskandar, George and et al.},
  journal={arXiv preprint arXiv:2506.22007},
  year={2025}
}

@article{chi2025wow,
  title={Wow: Towards a world omniscient world model through embodied interaction},
  author={Chi, Xiaowei and Jia, Peidong and Fan, Chun-Kai and Ju, Xiaozhu and Mi, Weishi and Zhang, Kevin and Qin, Zhiyuan and Tian, Wanxin and Ge, Kuangzhi and Li, Hao and others},
  journal={arXiv preprint arXiv:2509.22642},
  year={2025}
}

@article{liu2025towards,
  title={What matters in building vision--language--action models for generalist robots},
  author={Li, Xinghang and Li, Peiyan and Qian, Long and Liu, Minghuan and Wang, Dong and Liu, Jirong and Kang, Bingyi and Ma, Xiao and Kong, Tao and Zhang, Hanbo and Liu, Huaping},
  journal={Nature Machine Intelligence},
  year={2026},
  publisher={Nature Publishing Group},
  doi={10.1038/s42256-025-01168-7},
  url={https://doi.org/10.1038/s42256-025-01168-7}
}

@article{bjorck2025gr00t,
  title={Gr00t n1: An open foundation model for generalist humanoid robots},
  author={Bjorck, Johan and Casta{\~n}eda, Fernando and Cherniadev, Nikita and Da, Xingye and Ding, Runyu and Fan, Linxi and Fang, Yu and Fox, Dieter and Hu, Fengyuan and Huang, Spencer and others},
  journal={arXiv preprint arXiv:2503.14734},
  year={2025}
}

@article{li2023vision,
  title={Vision-language foundation models as effective robot imitators},
  author={Li, Xinghang and Liu, Minghuan and Zhang, Hanbo and Yu, Cunjun and Xu, Jie and Wu, Hongtao and Cheang, Chilam and Jing, Ya and Zhang, Weinan and Liu, Huaping and others},
  journal={arXiv preprint arXiv:2311.01378},
  year={2023}
}

@article{hancock2025actions,
  title={Actions as Language: Fine-Tuning VLMs into VLAs Without Catastrophic Forgetting},
  author={Hancock, Asher J. and Wu, Xindi and Zha, Lihan and Russakovsky, Olga and Majumdar, Anirudha},
  journal={arXiv preprint arXiv:2509.22195},
  year={2025}
}

@article{grover2025enhancing,
  title={Enhancing Generalization in Vision-Language-Action Models by Preserving Pretrained Representations},
  author={Grover, Shresth and Gopalkrishnan, Akshay and Ai, Bo and Christensen, Henrik I. and Su, Hao and Li, Xuanlin},
  journal={arXiv preprint arXiv:2509.11417},
  year={2025}
}

@article{cui2025emu3,
  title={Emu3. 5: Native multimodal models are world learners},
  author={Cui, Yufeng and Chen, Honghao and Deng, Haoge and Huang, Xu and Li, Xinghang and Liu, Jirong and Liu, Yang and Luo, Zhuoyan and Wang, Jinsheng and Wang, Wenxuan and others},
  journal={arXiv preprint arXiv:2510.26583},
  year={2025}
}

@article{huang2025vlmtdp,
  title={VLM-TDP: VLM-guided Trajectory-conditioned Diffusion Policy for Robust Long-Horizon Manipulation},
  author={Huang, Kefeng and Li, Tingguang and Liu, Yuzhen and Zhang, Zhe and Wang, Jiankun and Han, Lei},
  journal={arXiv preprint arXiv:2507.04524},
  year={2025}
}

@article{bu2025agibot,
  title={Agibot world colosseo: A large-scale manipulation platform for scalable and intelligent embodied systems},
  author={Bu, Qingwen and Cai, Jisong and Chen, Li and Cui, Xiuqi and Ding, Yan and Feng, Siyuan and Gao, Shenyuan and He, Xindong and Hu, Xuan and Huang, Xu and others},
  journal={arXiv preprint arXiv:2503.06669},
  year={2025}
}

@inproceedings{fu2024aloha,
  author       = {Zipeng Fu and
                  Tony Z. Zhao and
                  Chelsea Finn},
  editor       = {Pulkit Agrawal and
                  Oliver Kroemer and
                  Wolfram Burgard},
  title        = {Mobile {ALOHA:} Learning Bimanual Mobile Manipulation using Low-Cost
                  Whole-Body Teleoperation},
  booktitle    = {Conference on Robot Learning, 6-9 November 2024, Munich, Germany},
  series       = {Proceedings of Machine Learning Research},
  volume       = {270},
  pages        = {4066--4083},
  publisher    = {{PMLR}},
  year         = {2024},
  url          = {https://proceedings.mlr.press/v270/fu25b.html},
  bibsource    = {dblp computer science bibliography, https://dblp.org}
}

@article{douglas1973rdp,
  title={Algorithms for the reduction of the number of points required to represent a digitized line or its caricature},
  author={Douglas, David H and Peucker, Thomas K},
  journal={Cartographica: the international journal for geographic information and geovisualization},
  volume={10},
  number={2},
  pages={112--122},
  year={1973},
  publisher={University of Toronto Press}
}

@article{bai2025qwen2p5vl,
  title={Qwen2. 5-vl technical report},
  author={Bai, Shuai and Chen, Keqin and Liu, Xuejing and Wang, Jialin and Ge, Wenbin and Song, Sibo and Dang, Kai and Wang, Peng and Wang, Shijie and Tang, Jun and others},
  journal={arXiv preprint arXiv:2502.13923},
  year={2025}
}

@article{comanici2025gemini2p5,
  title={Gemini 2.5: Pushing the frontier with advanced reasoning, multimodality, long context, and next generation agentic capabilities},
  author={Comanici, Gheorghe and Bieber, Eric and Schaekermann, Mike and Pasupat, Ice and Sachdeva, Noveen and Dhillon, Inderjit and Blistein, Marcel and Ram, Ori and Zhang, Dan and Rosen, Evan and others},
  journal={arXiv preprint arXiv:2507.06261},
  year={2025}
}

@inproceedings{shi2025imagetokenizer,
  title={Scalable image tokenization with index backpropagation quantization},
  author={Shi, Fengyuan and Luo, Zhuoyan and Ge, Yixiao and Yang, Yujiu and Shan, Ying and Wang, Limin},
  booktitle={Proceedings of the IEEE/CVF International Conference on Computer Vision},
  pages={16037--16046},
  year={2025}
}

@article{bai2023qwen,
  title={Qwen technical report},
  author={Bai, Jinze and Bai, Shuai and Chu, Yunfei and Cui, Zeyu and Dang, Kai and Deng, Xiaodong and Fan, Yang and Ge, Wenbin and Han, Yu and Huang, Fei and others},
  journal={arXiv preprint arXiv:2309.16609},
  year={2023}
}

@article{yang2025qwen3,
  title={Qwen3 technical report},
  author={Yang, An and Li, Anfeng and Yang, Baosong and Zhang, Beichen and Hui, Binyuan and Zheng, Bo and Yu, Bowen and Gao, Chang and Huang, Chengen and Lv, Chenxu and others},
  journal={arXiv preprint arXiv:2505.09388},
  year={2025}
}

@article{shoeybi2019megatron,
  title={Megatron-lm: Training multi-billion parameter language models using model parallelism},
  author={Shoeybi, Mohammad and Patwary, Mostofa and Puri, Raul and LeGresley, Patrick and Casper, Jared and Catanzaro, Bryan},
  journal={arXiv preprint arXiv:1909.08053},
  year={2019}
}

@misc{mast3r_eccv24,
      title={Grounding Image Matching in 3D with MASt3R}, 
      author={Vincent Leroy and Yohann Cabon and Jerome Revaud},
      booktitle = {ECCV},
      year = {2024}
}

@article{black2024pi_0,
  title={A Vision-Language-Action Flow Model for General Robot Control},
  author={Black, Kevin and Brown, Noah and Driess, Danny and Esmail, Adnan and Equi, Michael and Finn, Chelsea and Fusai, Niccolo and Groom, Lachy and Hausman, Karol and Ichter, Brian and others},
  journal={arXiv preprint arXiv:2410.24164},
  year={2024}
}

@article{beyer2024paligemma,
  title={Paligemma: A versatile 3b vlm for transfer},
  author={Beyer, Lucas and Steiner, Andreas and Pinto, Andr{\'e} Susano and Kolesnikov, Alexander and Wang, Xiao and Salz, Daniel and Neumann, Maxim and Alabdulmohsin, Ibrahim and Tschannen, Michael and Bugliarello, Emanuele and others},
  journal={arXiv preprint arXiv:2407.07726},
  year={2024}
}

@inproceedings{zhai2023sigmoid,
  title={Sigmoid loss for language image pre-training},
  author={Zhai, Xiaohua and Mustafa, Basil and Kolesnikov, Alexander and Beyer, Lucas},
  booktitle={Proceedings of the IEEE/CVF international conference on computer vision},
  pages={11975--11986},
  year={2023}
}
\clearpage

\onecolumn
\appendices 
\renewcommand{\thefigure}{A.\arabic{figure}}
\renewcommand{\thetable}{A.\arabic{table}}
\setcounter{figure}{0}
\setcounter{table}{0}
\section{Implementation Details}\label{sec:app_impl_detail}
\subsection{World Model Planner}
\paragraph{Model Architecture and Tokenization} We construct the World Model with a total of 34.1 billion parameters, comprising 31.2 billion in the transformer layers and 2.9 billion in the embedding layers. For tokenization, we employ a unified vocabulary of 282,926 tokens. We utilize the pre-trained IBQ-Tokenizer from EMU3.5~\citep{cui2025emu3} for vision, which features a vocabulary size of 131,072 and quantizes each $16\times16$ patch into discrete tokens.
Particularly, in our setting, inputs are resized to a resolution of $512 \times 512$, resulting in 1024 visual tokens per image.
For language, we utilize the Qwen3 tokenizer~\citep{bai2023qwen} with a vocabulary size of 151,854.

\paragraph{Sequence Formatting} We set the max sequence length to 16,384, with each sequence formatted using interleaved text and visual tokens. For supervision, we provide only the instruction and the initial observations, and the model is required to predict the entire interleaved sequence auto-regressively. Since a sequence contains at most 16 images, we sample the timestamp within 1 second around a random goal image's timestamp, and task the model with predicting the subsequent steps. Therefore, the sequence could start at an arbitrary subtask stage, and would continue to generate the subtasks and goal images sequence until the task is finished or the stage number is reached as required in the prompt.
This sliding window style operation not only augments the dataset size and diversity, but also enables an endless closed-loop generation.

\paragraph{Model Training} For the training setup, we utilize Megatron-LM~\citep{shoeybi2019megatron} to train the world model, with the tensor-parallel size set to 8 and context-parallel size set to 1. The global batch size is 512, and the learning rate is $1\times10^{-5}$. We perform 2,000 steps of continued training on the open-source EMU3.5~\citep{cui2025emu3} checkpoint. 
We utilize 128 Nvidia H100 GPUs to train \our{} over a period of 2 days for post-training, employing a cosine learning rate scheduler with linear warmup. 

\subsection{Goal-conditioned VLA.}
\paragraph{Model Architecture}
We adopt a MoE-like architecture similar to \(\pi_0\)~\citep{black2024pi_0}, with PaliGemma-3B~\citep{beyer2024paligemma} as the backbone, and construct a 0.3B-scale action expert simultaneously. Block-wise causal masking is employed where the VLM block attends to its own features, the proprioception block (with weight sharing with the action block) attends to its own features and those of the VLM block, and the action block attends to all blocks; each individual block is fully bidirectional internally. In each inference step, the model takes six images as input, including three current observation images and three goal images, which are encoded into tokens via SigLIP~\citep{zhai2023sigmoid}. It also takes the current subtask prompt and proprioceptive signals representing the 6D pose of the robotic arm end-effector as inputs. Subsequently, a 10-step flow matching decoding process is performed to generate an action chunk with a length of 30 steps.

\paragraph{Two-Stage Training settings}
We adopt a two-stage training paradigm for GoalVLA. In the first stage, we leverage 200,000 trajectory samples from the AgiBot-Beta dataset~\citep{bu2025agibot}, utilizing annotated subtask text and goal images as additional conditioning. This stage consists of 100,000 training steps with a global batch size of 512 andd an initial learning rate of $5\times10^{-5}$. For the second stage, we fine-tune GoalVLA on 737 self-collected robot trajectories. We set a global batch size of 128 and a learning rate of $2\times10^{-5}$, training the model for 10 epochs (approximately 20k steps).

\paragraph{Inference Details}
In the real-robot deployment and testing phase, to mitigate the impact of motion errors and enable the model to adjust the robot with finer granularity, we only execute 10 out of the 30 steps in the action chunk inferred by the model. We adopt closed-loop absolute end-effector (EE) pose control to operate the robot arm. Since the model outputs delta EE poses, we calculate the absolute EE poses based on the currently read proprioceptive signals. For smoother control, we select the absolute EE poses at the 5th and 10th steps as target waypoints for execution.

\section{Dataset Construction and Visualization}\label{sec:app_dataset_construction}
\begin{table*}[!t]\tiny
\caption{Post-sampling dataset statistics for the embodied sub-task, goal image, and co-trained X2I datasets. M and B denote millions and billions, respectively.
}
\centering
\renewcommand{\arraystretch}{1.3}
\resizebox{0.8\textwidth}{!}{
\begin{tabular}{lccc}  
    \toprule  
    Data Type & Weight & Seq Num (M) & Token Num (B) \\  
    \midrule  
     Open-X Embodiment & 0.25 & 1.8 & 9.1  \\  
    AgiBot World Beta Single-View  & 0.075  & 0.16  & 2.2  \\  
    AgiBot World Beta Multi-View  & 0.075  & 0.32 & 3.8  \\  
    Self-collected Aloha Single-View  & 0.075  & 0.015  & 0.11  \\  
    Self-collected Aloha Multi-View  & 0.075  & 0.003  & 0.04  \\  
    Any-to-Image  & 0.45  & 3.3  & 15.2  \\  
    \bottomrule  
  \end{tabular}
\label{tab:data_stat}
\vspace{-15pt}
}
\end{table*}
With the proposed auto-divide and labeling framework, we totally split 1.2 million trajectories into interleaved subtask and goal image format, supporting 10 embodiments and multi-view generation.
For Agibot World Beta dataset, we further utilize Qwen2.5-VL 72B~\citep{bai2025qwen2p5vl} to refine the abstract task type to detailed instruction, and split the original skill description into fine-grained subtasks. The prompt we used in this framework is fully shown in the following code block.


\begin{figure*}[htbp]
    \centering
    \includegraphics[width=\linewidth]{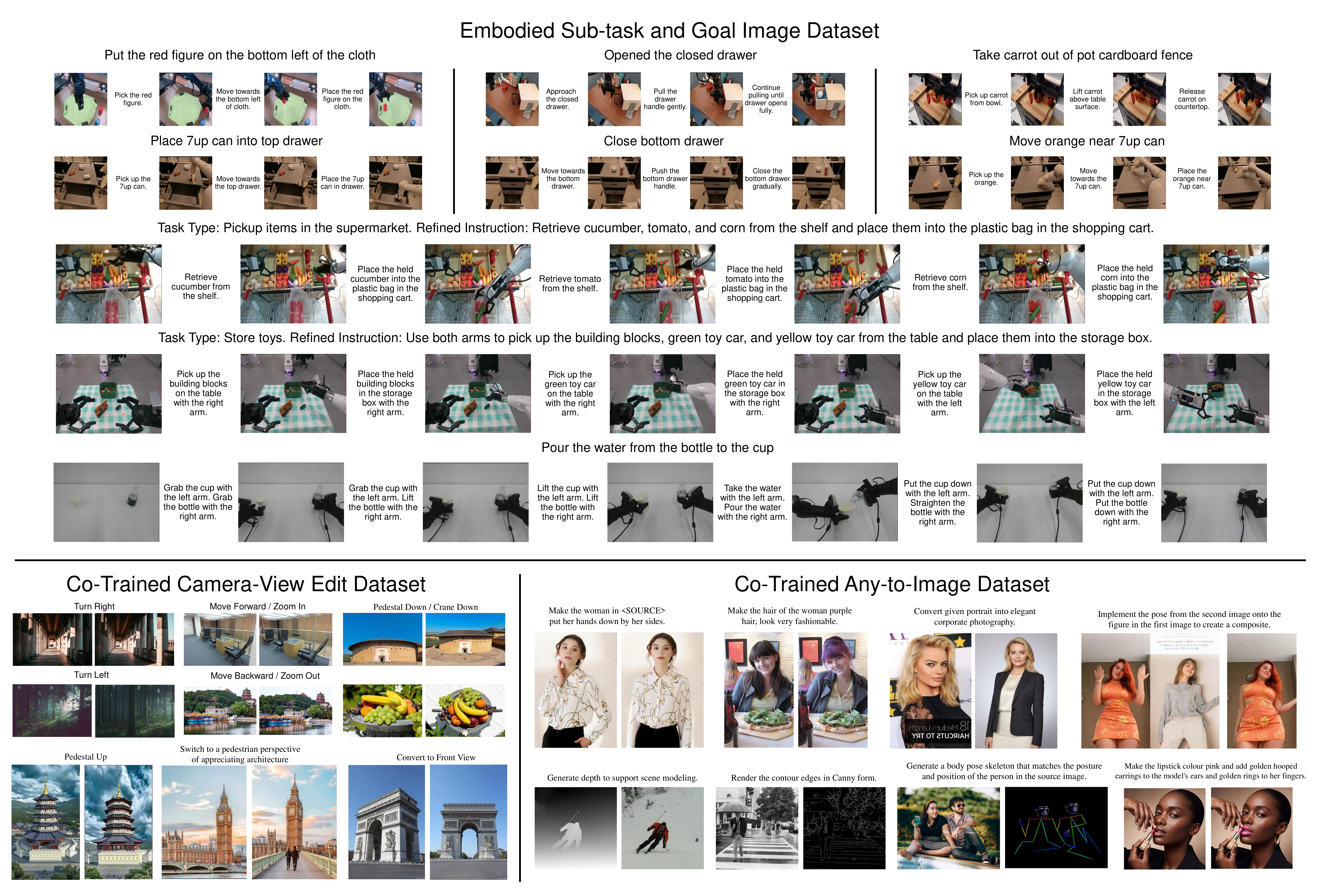}
    \caption{Visualization of samples from our constructed dataset. Our dataset consists of two main components: the Embodied Sub-task and Goal Image Dataset, and the Any-to-Image (X2I) dataset. In the bottom panel, we demonstrate the Camera-View Edit samples and other general Edit samples from the X2I dataset.}
    \label{fig:data_construction}
\end{figure*}

As a fundamental single-step generation task, X2I generation typically involves arbitrary multimodal interleaved inputs composed of text and any number of images, requiring the model to output a single image as the response. This imposes significant demands on the model's key capabilities, particularly in multimodal instruction following, maintaining subject/background consistency, adhering to world knowledge and rules, and controlling image style and texture. Mastering these challenging X2I capabilities will facilitate the model's robust evolution toward a more general Any-to-Any (X2X) generation paradigm, thereby advancing it into a more complex and powerful world model. To this end, we construct a large-scale X2I dataset (containing 15.2 billion tokens) for training to overcome the limitations in diversity, quality, and scale of existing open-source data. The in-house X2I dataset also integrates parts of multiple open-source datasets, including SEED-Data-Edit~\citep{ge2024seed}, WeatherStream~\citep{zhang2023weatherstream}, PromptFix~\citep{yu2024promptfix}, OmniGen-X2I~\citep{xiao2025omnigen}, ShareGPT-4o-Image~\citep{chen2025sharegpt}, ImgEdit~\citep{ye2025imgedit}, OmniGen2-X2I2~\citep{wu2025omnigen2}, MultiRef~\citep{Chen2025MultiRefCI}, and GPT-IMAGE-EDIT-1.5M~\citep{wang2025gpt}. To increase the spatial understanding and multi-view consistency, we further label the relative positions of multiple images shot from the same scene with mast3r~\citep{mast3r_eccv24}, and construct a camera-view edit dataset.

The prompt for subtask text segmentation and annotation:
\begin{lstlisting}

Your are an expert in robot indoor manipulation task analysis. The robot is exectuing the following instruction: {instruction}. Given {num_views} image sequence of the robot camera, you should analyze the subtask involved in the whole process.

The image sequences are organized as follows:
Image1_of_timestamp1, Image1_of_timestamp2, ..., Image1_of_timestamp{timestamp};
...
Image{num_views}_of_timestamp1, ..., Image{num_views}_of_timestamp{timestamp}

### NOTE:
1. The images are sampled uniformly, there could be multiple images belong to the same subtask.

2. The subtask should involved the manipulation skill and the manipulated object or target.

3. The subtask should be a sentence that is consistent with the instruction, and decribe the dyanmics between each adjacent images pair.

4. The subtask should be a single sentence, and the length should be less than 10 words and contains a single skill and the semantic description of the object or target.

5. All the skills should be involved in the skill library:
{skill_library}

6. Use predictive, grounded, and descriptive language that sounds like:
- "Approach the ..."
- "Close the gripper and pick up the ..."
- "Put down the ..."
- "Place the xxx onto the ..."
- "Close the ..."

7. The output should be a list of subtasks, each subtask is a single sentence.

8. The subtask number should ranged from 2 to 5.

9. The subtask should be continuous, and the from_timestamp and to_timestamp should be continuous.

10. The "from_timestamp" for the first subtask should be 1.

11. The "to_timestamp" for the last subtask should be {timestamp}.

12. The "from_timestamp" should be less than the "to_timestamp".

13. If the robot has two arms, add the utilized arm(s) information to the subtask (left/right/both).

### OUTPUT FORMAT:
```json
[
    {{"subtask": "subtask 1", "from_timestamp": 1, "to_timestamp": 2}},
    {{"subtask": "subtask 2", "from_timestamp": 2, "to_timestamp": 5}},
    ...
]
```

### Image Sequence {view_index} from {view_name} camera:
\end{lstlisting}

The prompt for instruction generation based on the task type and subtask description in AgiBot World Beta dataset:
\begin{lstlisting}
Your are an expert in robot indoor manipulation task analysis. The Dual-Arm Agibot is exectuing the following type of task: {task_type}. Given {num_views} image sequence of the robot camera and the subtask descriptions by temeral order, you should conclude the final instruction with a single brief sentence.
    
The image sequences are organized as follows:
Image1_of_timestamp1, Image1_of_timestamp2, ..., Image1_of_timestamp{timestamp};
...
Image{num_views}_of_timestamp1, ..., Image{num_views}_of_timestamp{timestamp}

### Subtask Descriptions:
{subtask_descriptions}
    
### Image Sequence {view_index} from {view_name} camera:
\end{lstlisting}

\section{Visualization of Generated Results for Unconstrained Scenarios}\label{sec:app_vis_unconstrained}

\begin{figure*}[!h]
    \centering
    \includegraphics[width=0.98\linewidth]{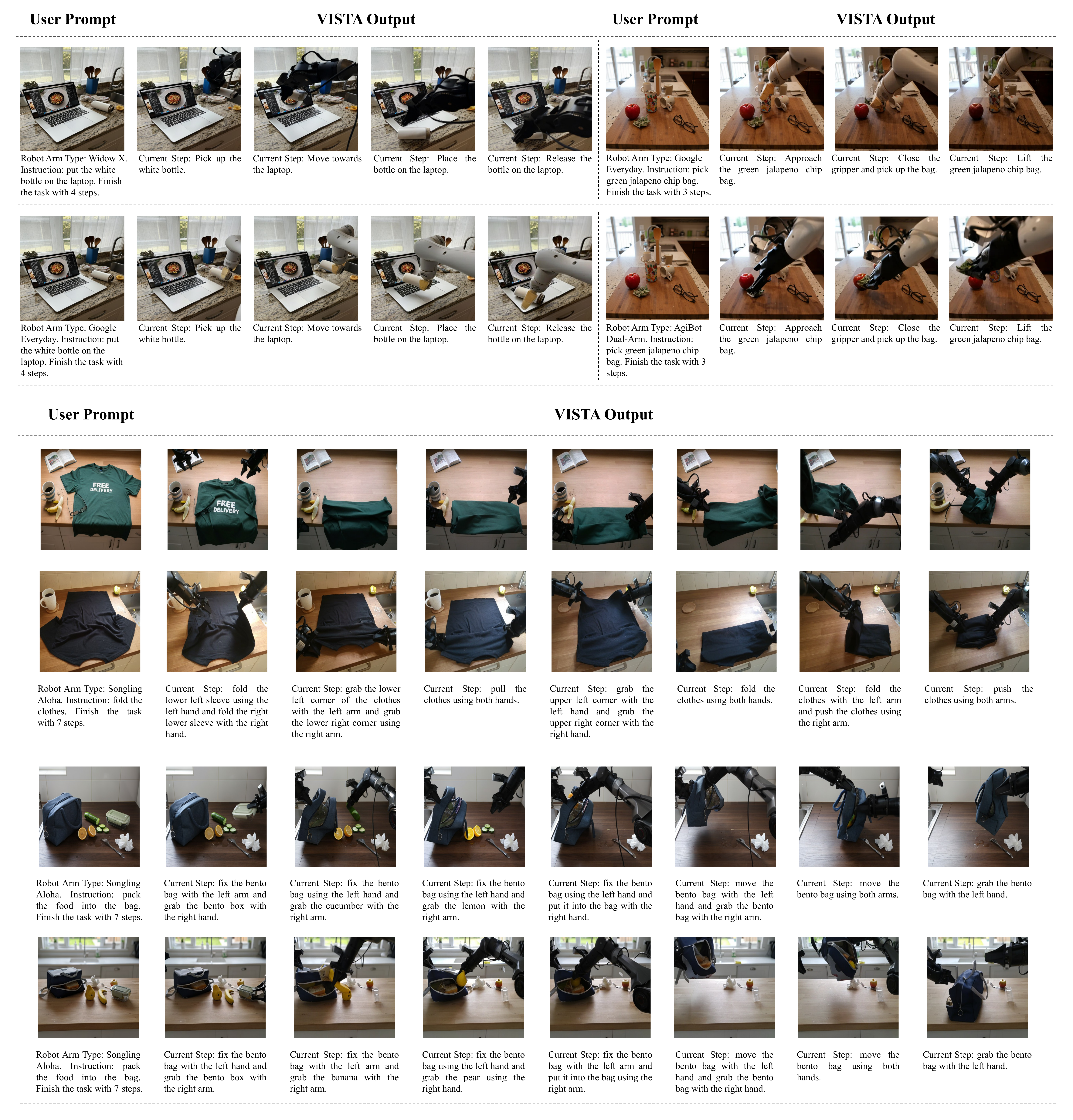}
    \caption{Generated samples of \our{} on diverse unconstrained scenarios. These scenarios are dramatically different from the training sample in terms of layout, object appearance, background, and camera view.}
    \label{fig:banana_results_part2}
\end{figure*}

\begin{figure*}[!h]
    \centering
    \includegraphics[width=0.98\linewidth]{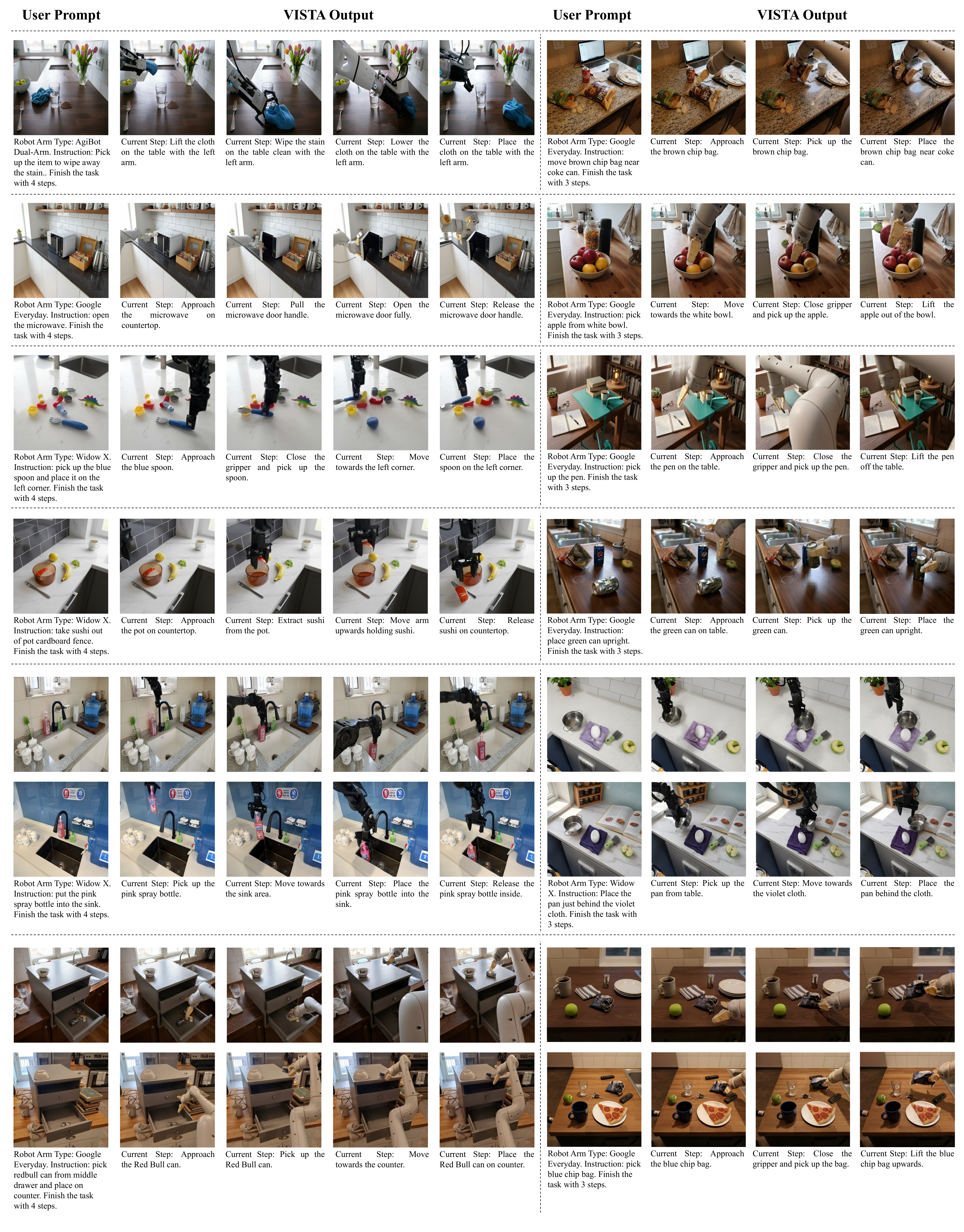}
    \caption{More visualization results for the generated interleaved sequences on unconstrained scenarios.}
    \label{fig:banana_results_part1}
\end{figure*}

We evaluate the generative performance of our world model across diverse, unconstrained scenarios, including images synthesized by text-to-image models (e.g., Nano Banana~\citep{comanici2025gemini2p5}) and real-world photographs captured via mobile devices. The results are shown in Fig.~\ref{fig:banana_results_part2} and Fig.~\ref{fig:banana_results_part1}. On these novel scenarios, \our{} can alternately generate reasonable subtasks and physically plausible, instruction-following goal images according to the specified robot arm type and task. For long-horizon tasks such as folding clothes and tidying desktops, \our{} is also capable of decomposing tasks into key steps and generating informative keyframes that provide effective guidance.

\section{Illustration of Real-World Experiment Setup}\label{sec:app_experiment_setup}
\begin{figure*}[!h]
    \centering
    \includegraphics[width=1\linewidth]{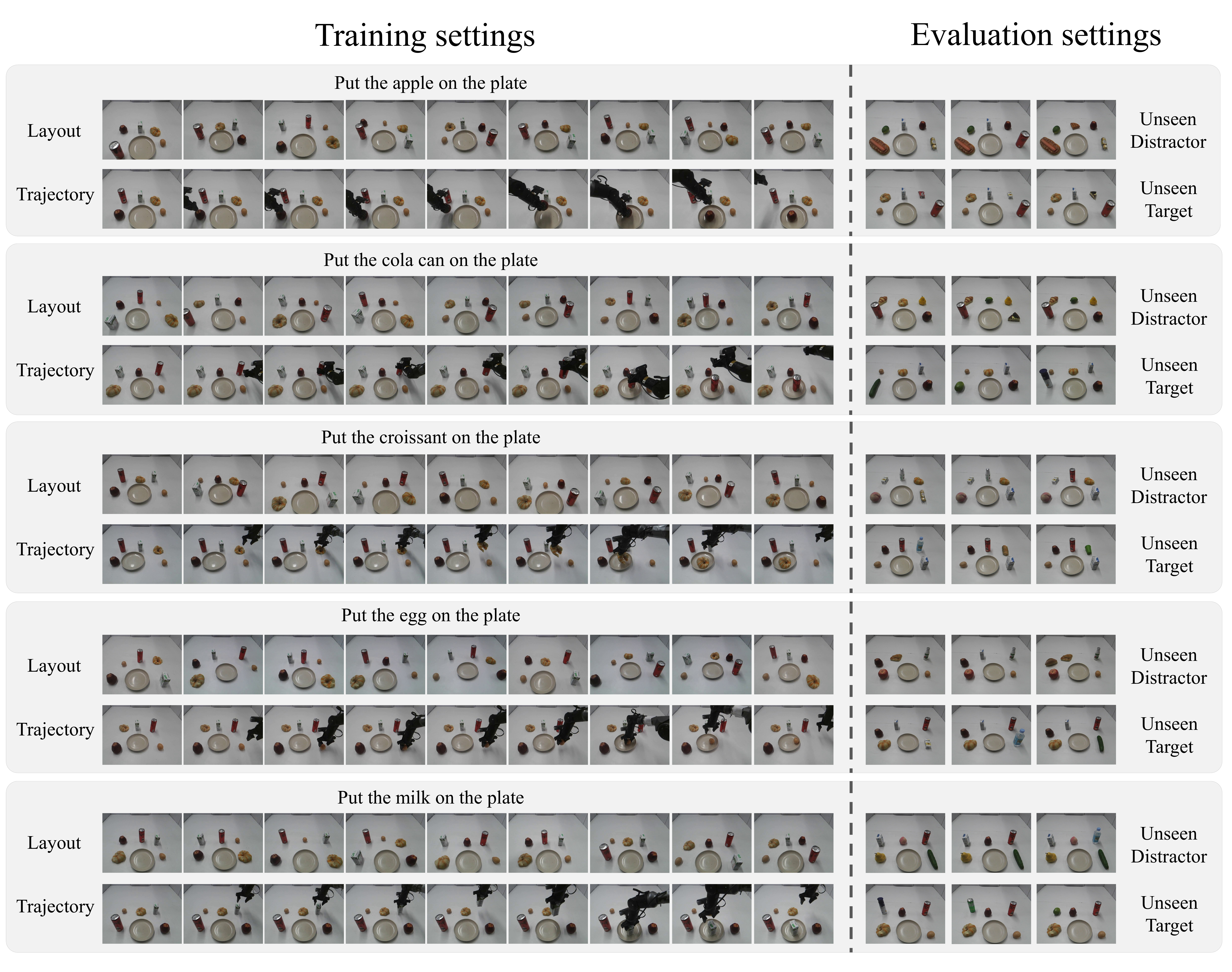}
    \vspace{-6pt}
    \caption{Visualization of training dataset and in-domain scenarios setup.}
    \vspace{-15pt} 
    \label{fig:scenario_settings}
\end{figure*}

We present the setup for in-domain scenarios in Fig.~\ref{fig:scenario_settings}. The distribution of our training data is shown on the left side of Fig.~\ref{fig:scenario_settings}. The training scenes contain five fixed objects, and the training set consists of five tasks (i.e., put one of the objects on the plate). Each task includes approximately 150 trajectories, amounting to a total of 737 trajectories. It can be seen that our training data is highly homogeneous in both object placement and scene composition. On the right side of Fig.~\ref{fig:scenario_settings}, we show the settings for unseen distractors and unseen targets used in our testing. For the unseen distractor setting, we first align with the object placement layouts present in the training set, then replace \(1–3\) of the distractors with unseen objects. The position of the target object remains unchanged in this setting, which thus primarily evaluates the model’s robustness to distractor objects. For the unseen target setting, we also align with the object placement layouts from the training set, then replace the target object with an unseen one. The position of the target object is also fixed in this setting, which mainly assesses the model’s generalization capability for the grounding and grasping of unseen objects.

\begin{figure*}[!h]
    \centering
    \includegraphics[width=0.9\linewidth]{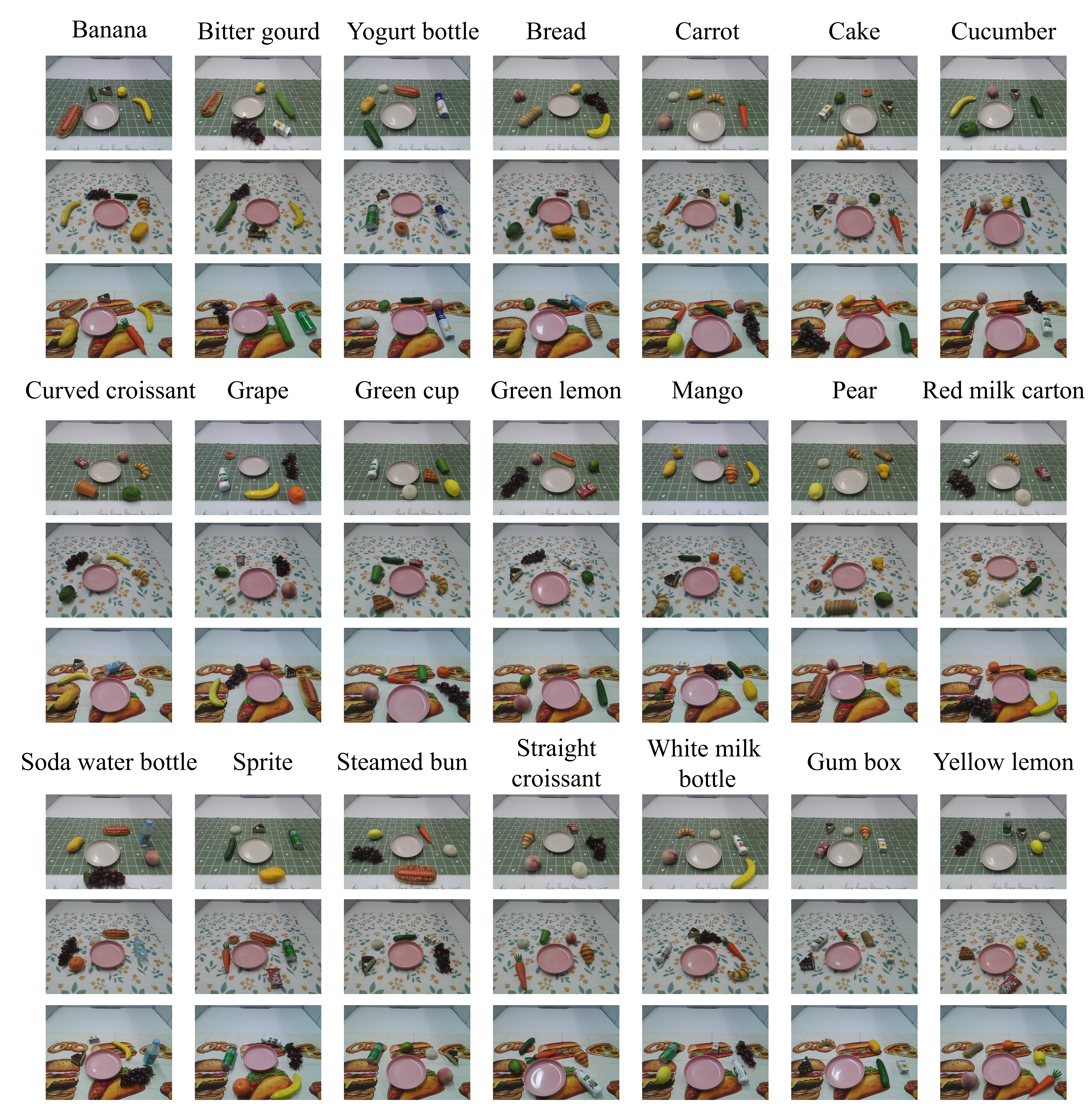}
    \vspace{-6pt}
    \caption{Visualization of novel scenarios setup.}
    \label{fig:novel_scenario_vis}
\end{figure*}

Fig.~\ref{fig:novel_scenario_vis} presents the setup for novel scenarios, which includes 21 unseen objects, tablecloths with three distinct patterns, and plates in unseen colors, forming a total of 63 novel scenarios for evaluation. Given the lack of diversity in the placement positions of objects within the training data, we consistently place the target object within the reachable and graspable range of the robotic arm in the setup of novel scenarios, while introducing randomness in its specific position to ensure a distinction from the training set.

\section{Visualization of generated goal images for novel scenarios}\label{sec:app_novel_scenario}
\begin{figure*}[!h]
    \centering
    \includegraphics[width=0.9\linewidth]{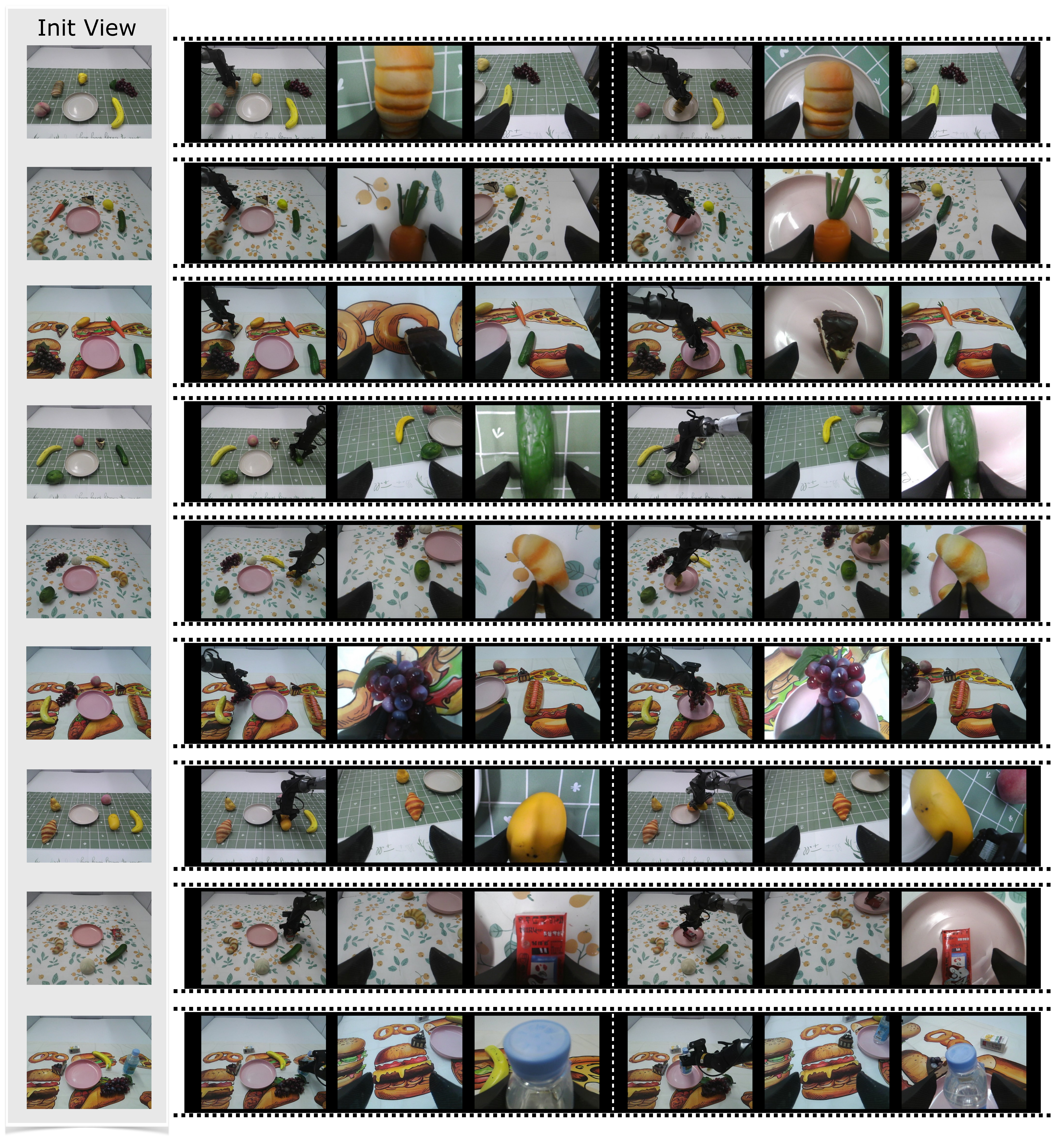}
    \caption{Visualization of generated goal images for novel scenarios.}
    \label{fig:supp_mixture_gen}
\end{figure*}

We present more goal images generated by \our{} on novel scenarios in Fig.~\ref{fig:supp_mixture_gen}. It can be observed that the generated goal images are almost all timing-accurate, effectively capturing the critical moments of picking and placing, and thus providing precise spatial positional guidance for object grasping and placement. Benefiting from the powerful image generation capability of Emu3.5, \our{} is able to maintain high consistency in both background and object appearance even when confronted with previously unseen scenarios. Furthermore, after fine-tuning on a large-scale cross-embodiment robotic dataset, \our{} demonstrates remarkable multi-view consistent generation. As illustrated in the fourth row of Fig.~\ref{fig:supp_mixture_gen}, in the generated images of cucumber placement, the cucumber in the plate and the gripper of the right arm are clearly visible from the wrist camera of the left arm, with their relative spatial positions being almost perfectly consistent. This strong multi-view spatial consistency enables \our{} to provide clear and reliable spatial relationship cues for the goal-conditioned VLA, thereby enhancing its ability to perceive the spatial positions of objects.

In addition, we find that our model is capable of generating corresponding reasonable grasping modes for different object placement configurations. As shown in the last row of Fig.~\ref{fig:supp_mixture_gen}, when generating goal images for grasping a soda water bottle, the gripper adopts a horizontal, forward grasp rather than the vertical downward grasp commonly observed for other objects. This ability to synthesize diverse and object-specific grasping strategies is crucial for handling a wide range of unseen objects. With more robotic manipulation datasets, our model is promising to generate reasonable and critical goal images for a wide range of complex manipulation tasks.

\clearpage
\section{Emerging Capability Analysis}\label{sec:app_emerging_capability}
\begin{figure*}[!h]
    \centering
    \includegraphics[width=1\linewidth]{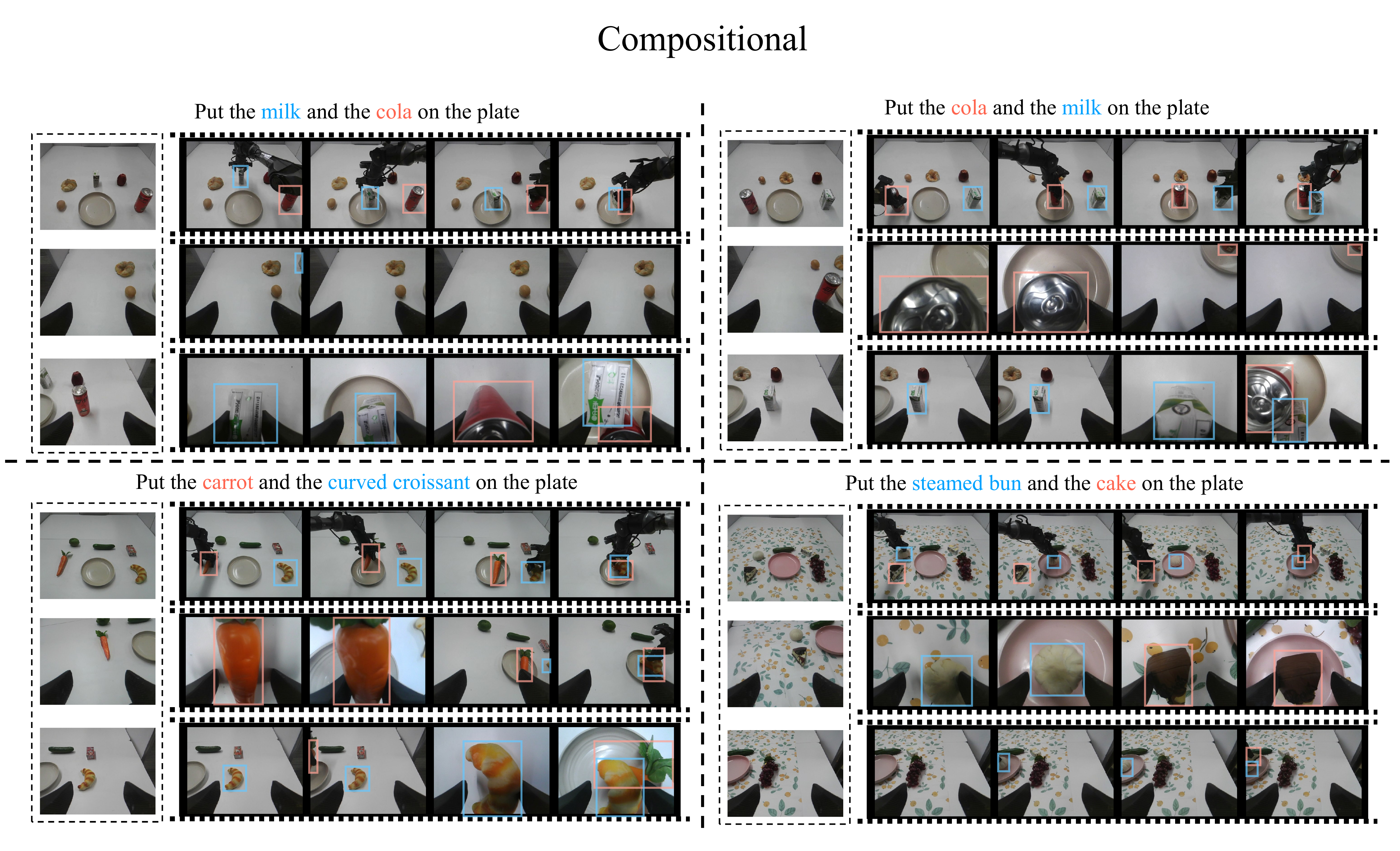}
    \caption{Visualization of \our{} generated sequences on compositional tasks.}
    \label{fig:supp_comp}
\end{figure*}

\begin{figure*}[!h]
    \centering
    \includegraphics[width=1\linewidth]{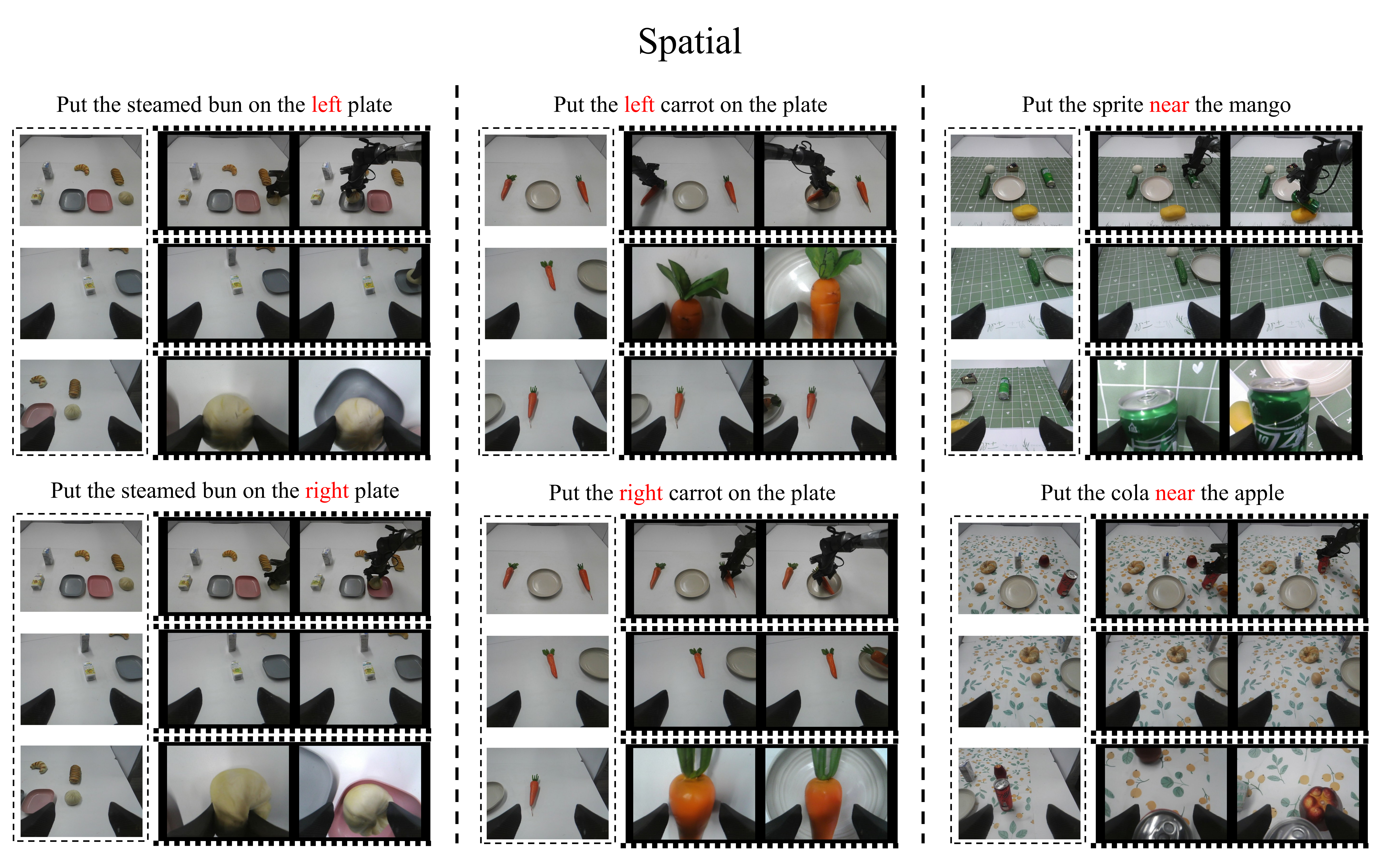}
    \caption{Visualization of \our{} generated sequences on spatial understanding tasks.}
    \label{fig:supp_spatial}
\end{figure*}

\begin{figure*}[!h]
    \centering
    \includegraphics[width=1\linewidth]{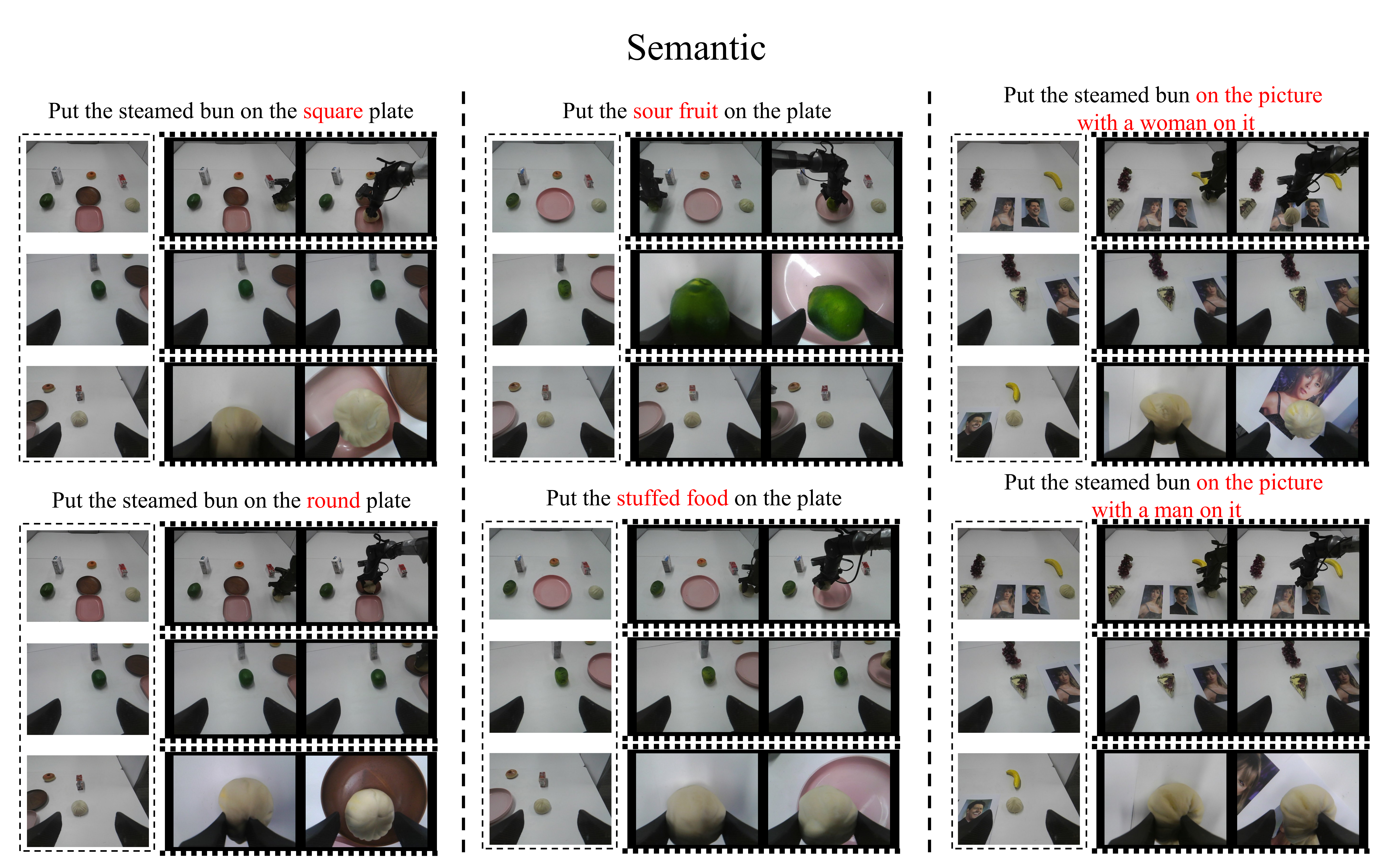}
    \caption{Visualization of \our{} generated sequences on semantic understanding tasks.}
    \label{fig:supp_semantic}
\end{figure*}

We further investigate the generalization boundaries of \our{} in generating goal images. Fig.~\ref{fig:supp_comp},~\ref{fig:supp_spatial} and ~\ref{fig:supp_semantic} presents qualitative visualizations of compositional tasks, spatial understanding tasks, and semantic understanding tasks generated by \our{}. 

For compositional tasks, \our{} is able to accurately interpret the given instructions and sequentially generate goal images for manipulating multiple objects, while maintaining spatiotemporal consistency in object positions, as shown in Fig.~\ref{fig:supp_comp}. These results suggest that our approach has the potential to scale to more complex long-horizon manipulation tasks and to support flexible composition of subtasks. Moreover, such compositional visual guidance enables goal-conditioned VLAs to achieve controllable composition of manipulation skills. 

For spatial understanding tasks, \our{} demonstrates the ability to comprehend instructions involving simple spatial relationships. As shown in Fig.~\ref{fig:supp_spatial}, \our{} correctly interprets spatial descriptions of both placement targets and grasping targets. In addition, we observe that \our{} can generalize to novel task types, such as “put the Sprite near the mango”, in which the generated goal image shows the robot placing the object outside the plate in accordance with the instruction.

For semantic understanding tasks, \our{} similarly identifies the manipulated objects and placement targets based on semantic cues and generates the corresponding goal images. The Fig.~\ref{fig:supp_semantic} illustrates several representative examples, including instructions describing the shape and color of the plate, as well as more challenging cases that require semantic understanding of pictures displayed on the tabletop.

We attribute these instruction-following capabilities to the strong text-to-image generation and editing abilities of Emu3.5. After fine-tuning on robotic datasets, we find that a portion of this instruction-following ability is retained, although it remains limited and typically exhibits hallucinations. We believe that with more diverse robotic data, \our{} has significant potential to further improve its generation quality on these tasks. 

\begin{figure*}[!h]
    \centering
    \includegraphics[width=0.95\linewidth]{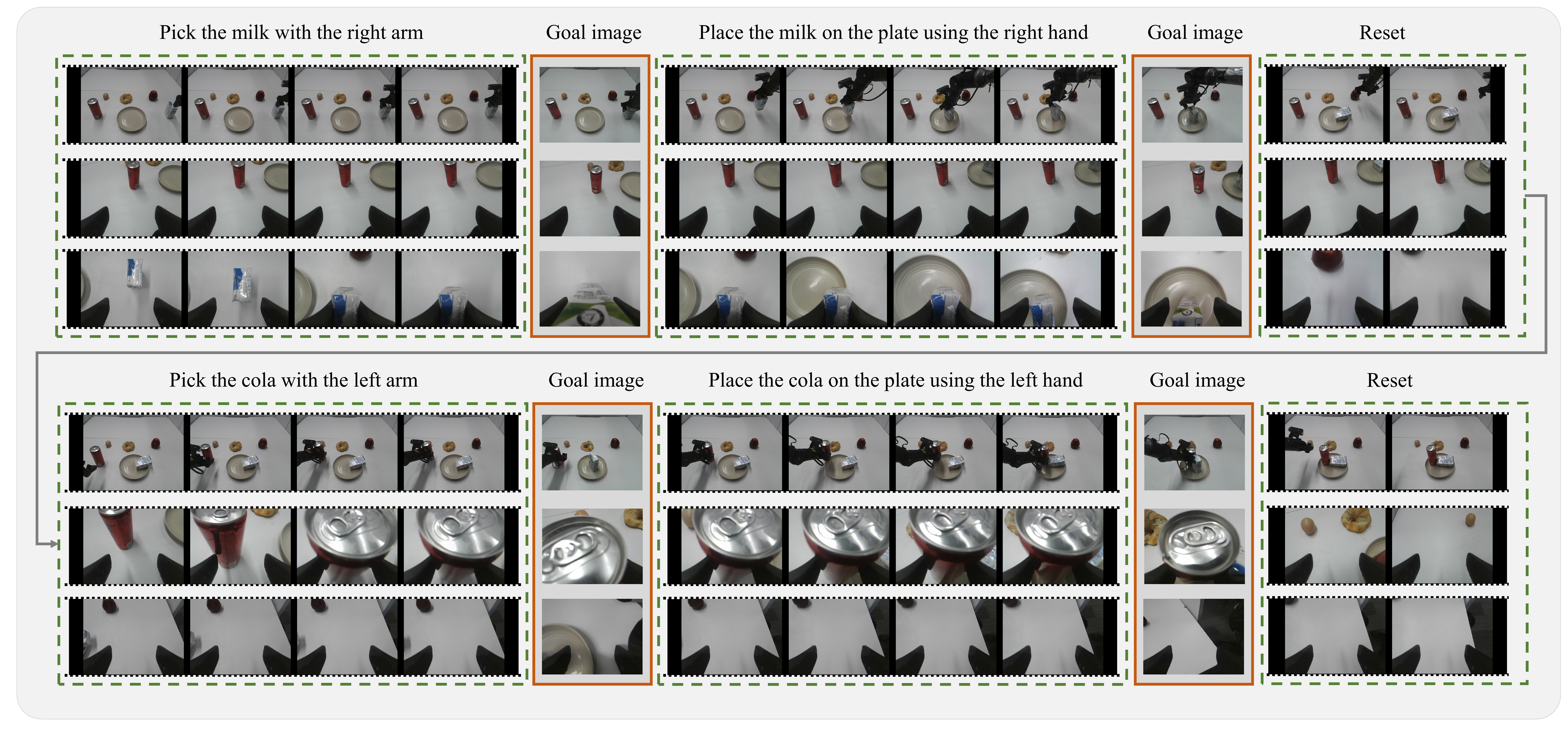}
    \caption{Visualization of goal-conditioned execution trajectory for compositional task. For the reset stage, our GoalVLA is trained via language instructions without requiring goal images.}
    \vspace{-15pt} 
    \label{fig:supp_emerging_execute_comp}
\end{figure*}

\begin{figure*}[!h]
    \centering
    \includegraphics[width=0.8\linewidth]{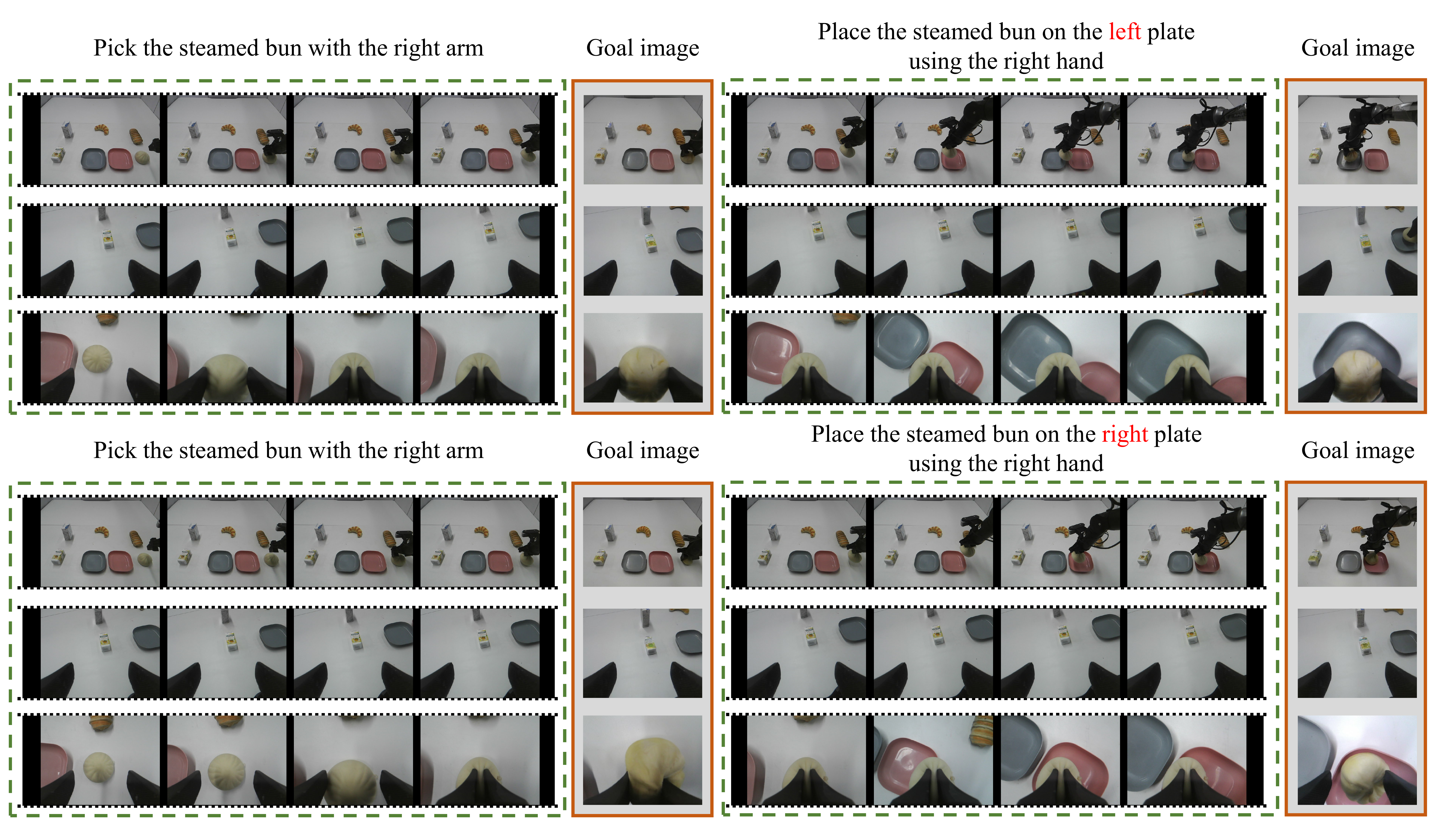}
    \vspace{-6pt}
    \caption{Visualization of goal-conditioned execution trajectories for spatial understanding tasks.}
    \vspace{-15pt} 
    \label{fig:supp_emerging_execute_spatial}
\end{figure*}

\begin{figure*}[!h]
    \centering
    \includegraphics[width=0.8\linewidth]{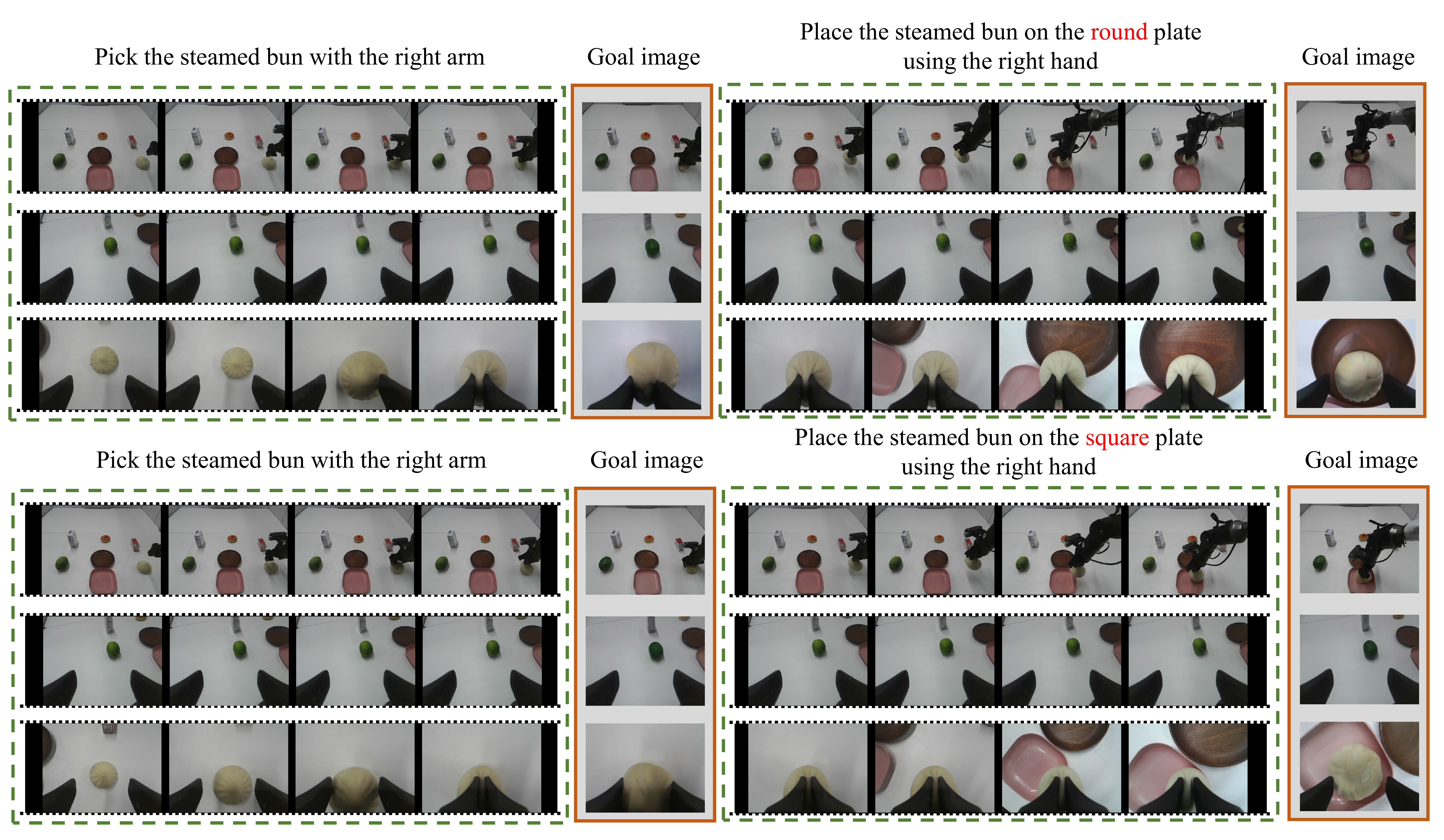}
    \vspace{-6pt}
    \caption{Visualization of goal-conditioned execution trajectories for semantic understanding tasks.}
    \vspace{-15pt} 
    \label{fig:supp_emerging_execute_semantic}
\end{figure*}

By leveraging the instruction understanding capability of the world model, the GoalVLA can focus more on action generation and goal image following, thereby alleviating its limitations in instruction comprehension. We present several real-world execution examples in Fig.~\ref{fig:supp_emerging_execute_comp},~\ref{fig:supp_emerging_execute_spatial} and~\ref{fig:supp_emerging_execute_semantic}. As shown, by integrating the world model with stronger instruction understanding, the manipulation capacity of the VLA can be further raised, enabling not only flexible composition of manipulation skills but also execution of tasks that require instruction-level reasoning.

\clearpage
\section{Visualization and Analysis of Execution Results}\label{sec:app_execution_result}

\begin{figure*}[!h]
    \centering
    \includegraphics[width=0.98\linewidth]{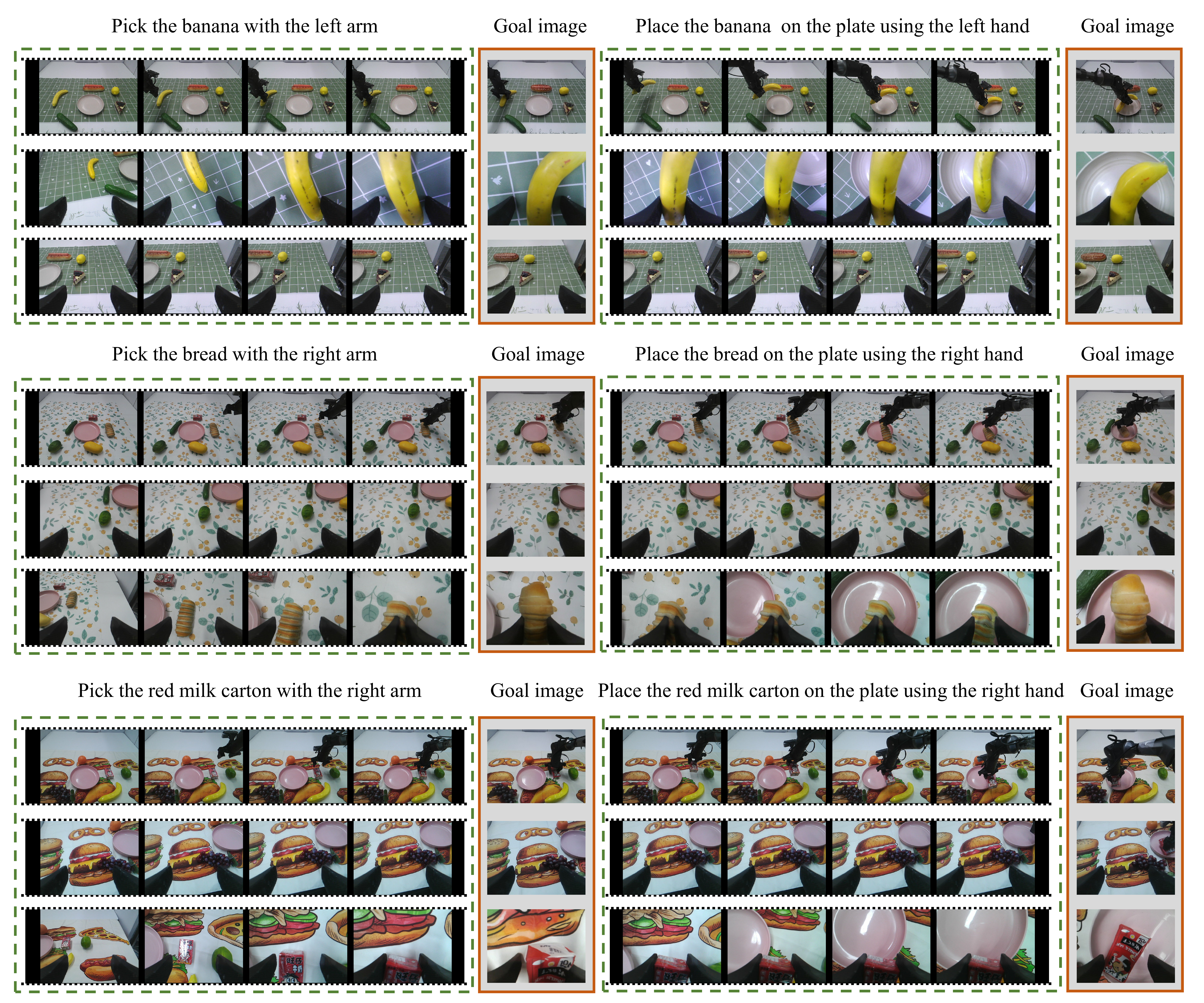}
    \caption{Visualization of goal-conditioned execution trajectories.}
    \label{fig:execute_visualize}
\end{figure*}

We visualize additional execution results, as shown in Fig.~\ref{fig:execute_visualize}. By comparing the final frame of each subtask stage with its corresponding goal image, we observe that the robot arm posture largely aligns with that shown in the goal image. This indicates that GoalVLA can accurately capture the visual features related to the robot’s spatial configuration provided by the goal images and then robustly generate the corresponding action chunks to reach the specified positions and execute the desired manipulations, even in novel scenarios with significant visual distractions.

\begin{figure*}[!h]
    \centering
    \includegraphics[width=0.98\linewidth]{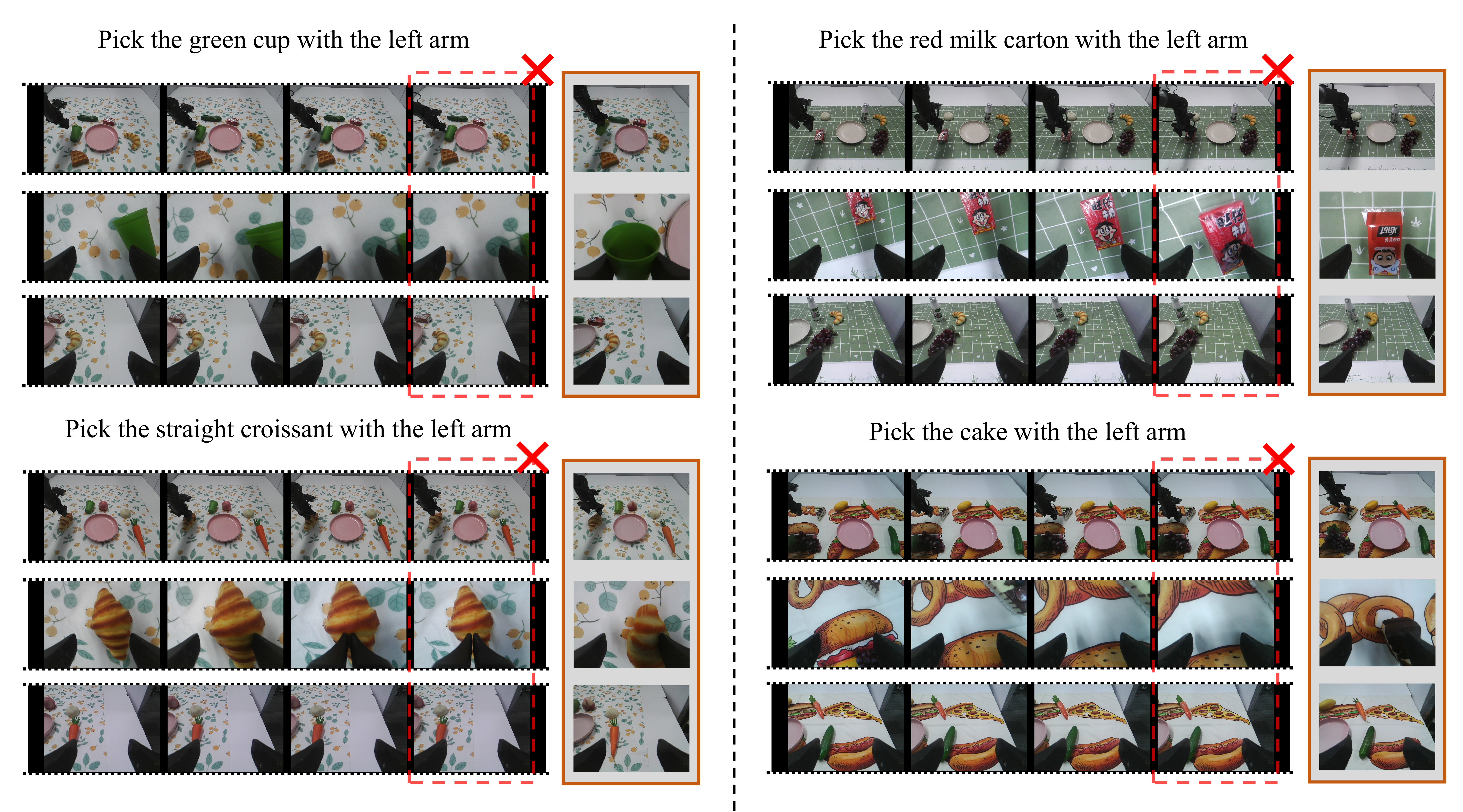}
    \caption{Visualization of failed execution trajectories. The left side shows failures caused by inaccurate timing of the generated goal images, and the right side shows failures caused by spatial misalignment of the generated goal images.}
    \label{fig:failure_case}
\end{figure*}

We also analyze several failure cases observed during execution. We identify two categories of failures caused by suboptimal goal image generation quality. The first category arises from inaccurate timing of the generated goal images, where the images do not precisely correspond to the critical moments of picking or placing. As illustrated in the bottom-left example of Fig.~\ref{fig:failure_case}, the goal image is generated at a moment prior to object grasping, when the robot arm has not yet descended sufficiently and the gripper has not fully enclosed the object. During execution, GoalVLA overly focuses on matching the robot pose depicted in the goal image, resulting in insufficient downward motion and eventual grasp failure. This suggests that GoalVLA relies heavily on accurate grasping position cues provided by the goal images, while exhibiting limited capability to adapt to the actual execution state observed online. The top-left example in Fig.~\ref{fig:failure_case} shows another failure case where the goal image corresponds to a moment after the object has already been grasped and slightly lifted. In this case, the goal image fails to provide precise spatial guidance for the grasping phase, leading to positional deviations and grasp failure. To mitigate failures caused by imprecise goal image timing, we employ a random goal image offset during the training of GoalVLA, which can partially alleviate this issue. However, in certain cases, GoalVLA still exhibits suboptimal performance. We leave further improvements on this aspect to future work.

The second category of failures stems from spatial misalignment in the generated goal images. As shown in the top-right example of Fig.~\ref{fig:failure_case}, the goal image from the left wrist camera is not fully aligned with the target object and is slightly shifted to the left, causing the left gripper to collide with the object during execution and resulting in grasp failure. Similarly, in the bottom-right example of Fig.~\ref{fig:failure_case}, the goal images from the left wrist camera exhibit a comparable leftward offset, again leading to execution failure. In such cases, GoalVLA is required to balance guidance from the goal images with the current visual observations, rather than relying exclusively on the goal images to generate actions. Achieving this balance calls for a more refined network design, which we plan to explore in future work.

Finally, limited by the diversity of the training data, GoalVLA struggles to accurately follow goal images that specify target positions significantly outside the training distribution, even when the goal images themselves are accurately generated. We believe that increasing the spatial diversity of training data is crucial for improving GoalVLA’s generalization, and we will explore training with more diverse datasets in future work.

\clearpage

\end{document}